\documentclass[letterpaper]{article} 
\usepackage{aaai25}  
\usepackage{times}  
\usepackage{helvet}  
\usepackage{courier}  
\usepackage[hyphens]{url}  
\usepackage{graphicx} 
\usepackage{natbib}  
\usepackage{caption} 
\usepackage{algorithm}
\usepackage{algorithmic}
\usepackage{amsmath}
\usepackage{amssymb}
\usepackage{caption}
\usepackage{subcaption}
\usepackage{xspace}

\DeclareMathOperator{\E}{\mathbb{E}}

\newcommand{\libname}{\textsf{Virny}\xspace}

\newcommand*{\diabetes}{\texttt{diabetes}\xspace}
\newcommand*{\german}{\texttt{german}\xspace}
\newcommand*{\folk}{\texttt{folk-income}\xspace}
\newcommand*{\law}{\texttt{law-school}\xspace}
\newcommand*{\bank}{\texttt{bank}\xspace}

\newcommand*{\folkcov}{\texttt{folk-pubcov}\xspace}
\newcommand*{\student}{\texttt{student-performance}\xspace}

\usepackage{newfloat}
\usepackage{listings}
\DeclareCaptionStyle{ruled}{labelfont=normalfont,labelsep=colon,strut=off} 
\floatstyle{ruled}
\newfloat{listing}{tb}{lst}{}
\floatname{listing}{Listing}
\title{An Epistemic and Aleatoric Decomposition of Arbitrariness to Constrain the Set of Good Models}
\author {
    Falaah Arif Khan\textsuperscript{\rm 1},
    Denys Herasymuk \textsuperscript{\rm 2},
    Nazar Protsiv\textsuperscript{\rm 2},
    Julia Stoyanovich\textsuperscript{\rm 1}
}
\affiliations {
    \textsuperscript{\rm 1} New York University
    \textsuperscript{\rm 2} Ukrainian Catholic University\\
    fa2161@nyu.edu, herasymuk@ucu.edu.ua, protsiv.pn@ucu.edu.ua, stoyanovich@nyu.edu
}

\begin{document}

\maketitle

\begin{abstract}
Recent research reveals that machine learning (ML) models are highly sensitive to minor changes in their training procedure, such as the inclusion or exclusion of a single data point, leading to conflicting predictions on individual data points; a property termed as arbitrariness or instability in ML pipelines in prior work. Drawing from the uncertainty literature, we show that stability decomposes into epistemic and aleatoric components, capturing the consistency and confidence in prediction, respectively. We use this decomposition to provide two main contributions.

Our first contribution is an extensive empirical evaluation. We find that (i) epistemic instability can be reduced with more training data whereas aleatoric instability cannot; (ii) state-of-the-art ML models have aleatoric instability as high as 79\% and aleatoric instability disparities among demographic groups as high as 29\% in popular fairness benchmarks; and (iii) fairness pre-processing interventions generally increase aleatoric instability more than in-processing interventions, and both epistemic and aleatoric instability are highly sensitive to data-processing interventions and model architecture. 

Our second contribution is a practical solution to the problem of systematic arbitrariness.  We propose a model selection procedure that includes epistemic and aleatoric criteria alongside existing accuracy and fairness criteria, and show that it successfully narrows down a large set of good models (50-100 on our datasets) to a handful of stable, fair and accurate ones. We built and publicly released a python library to measure epistemic and aleatoric multiplicity in any ML pipeline alongside existing confusion-matrix-based metrics, providing practitioners with a rich suite of evaluation metrics to use to define a more precise criterion during model selection.

\end{abstract}

%

\section{Introduction}

Recent research has shown that ML models are highly sensitive to small perturbations in their training pipelines. For example, the inclusion or exclusion of a small set of data points can change which model is learned, which can lead to conflicting predictions on individual data points~\cite{black_leave_one_out_Fairness, black_multiplicity_22}. This has been studied under several names, including the Rashomon Effect~\cite{breiman_two_cultures, paes2023inevitability}, model multiplicity~\cite{marx2020predictive, black_multiplicity_22}, dataset multiplicity~\cite{meyer2023dataset}, arbitrariness in predictions~\cite{cooper2023arbitrariness, long2023arbitrariness}, and instability in ML pipelines~\cite{khan2023fairness, subbaswamy2021evaluating, liu2022jitter}. Multiplicity is the phenomenon where models with equivalent accuracy (usually the primary criterion for model selection) differ in individual predictions (predictive multiplicity) or in their internals (procedural multiplicity). Instability or arbitrariness can be understood as the magnitude of multiplicity that a training procedure is likely to admit: a more unstable model has higher predictive multiplicity. Put differently, predictive multiplicity in a set of viable models is equivalent to instability within a single model\footnote{We use `arbitrariness' in predictions, `instability' in a single model and 'multiplicity' in a set of viable models interchangeably.}. 

\begin{table}[t]
\centering
\caption{Illustrative Example: $m_1$, $m_2$, $m_3$ are models with equivalent accuracy, $x_1$, $x_2$, $x_3$ are test data points}
\begin{tabular}{ccccc}
 & $m_1$ & $m_2$ & $m_3$ &  \\
$x_1$ & 0.9 & 0.9 & 0.9 & stable  \\
$x_2$ & 0.2 & 0.85 & 0.9 &  unstable (inconsistent) \\
$x_3$ & 0.49 & 0.51 & 0.52 &  unstable (under-confident)\\
\end{tabular}

\label{tab:example}
\end{table}

\begin{figure*}[t!]
     \centering
     \begin{subfigure}[b]{0.635\textwidth}
         \centering
         \includegraphics[width=\textwidth]{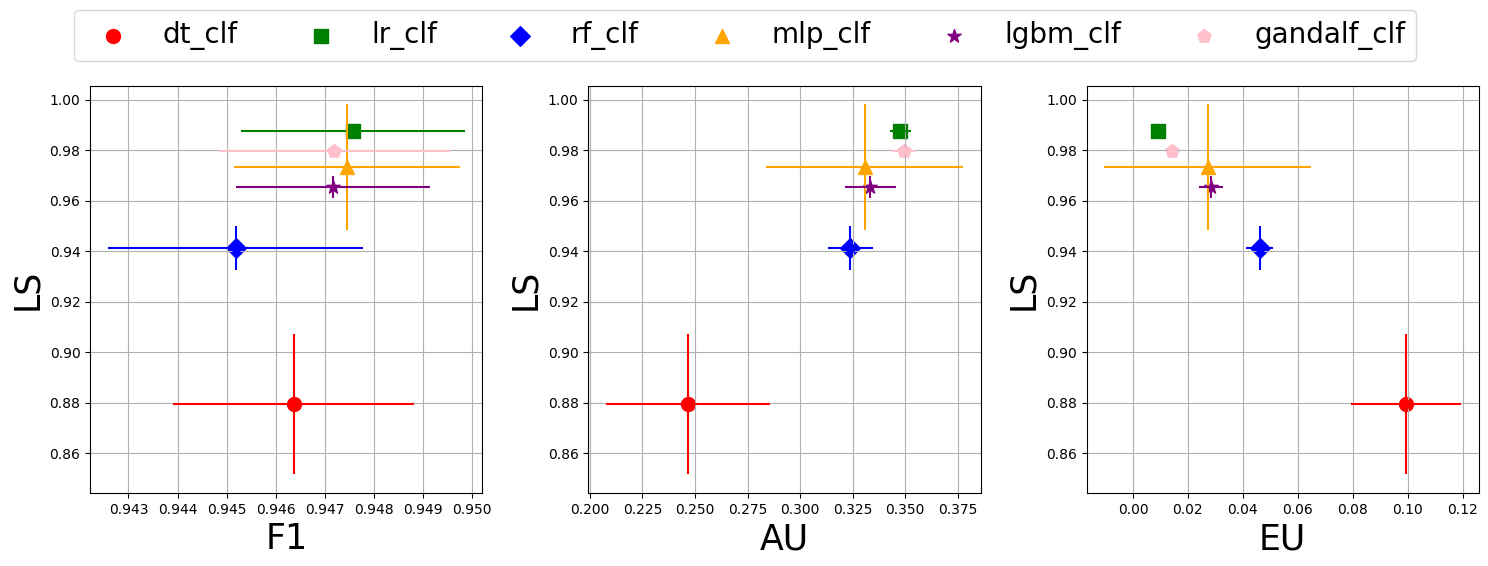}
         \caption{Stability decomposition: Label stability (LS) and F1 are somewhat orthogonal, and models with comparable LS have different epistemic uncertainty (EU) and aleatoric uncertainty (AU).}
         \label{fig:law-metrics}
     \end{subfigure}
     \hspace{10pt}
     \hfill
     \begin{subfigure}[b]{0.33\textwidth}
         \centering
         \includegraphics[width=0.8\textwidth]{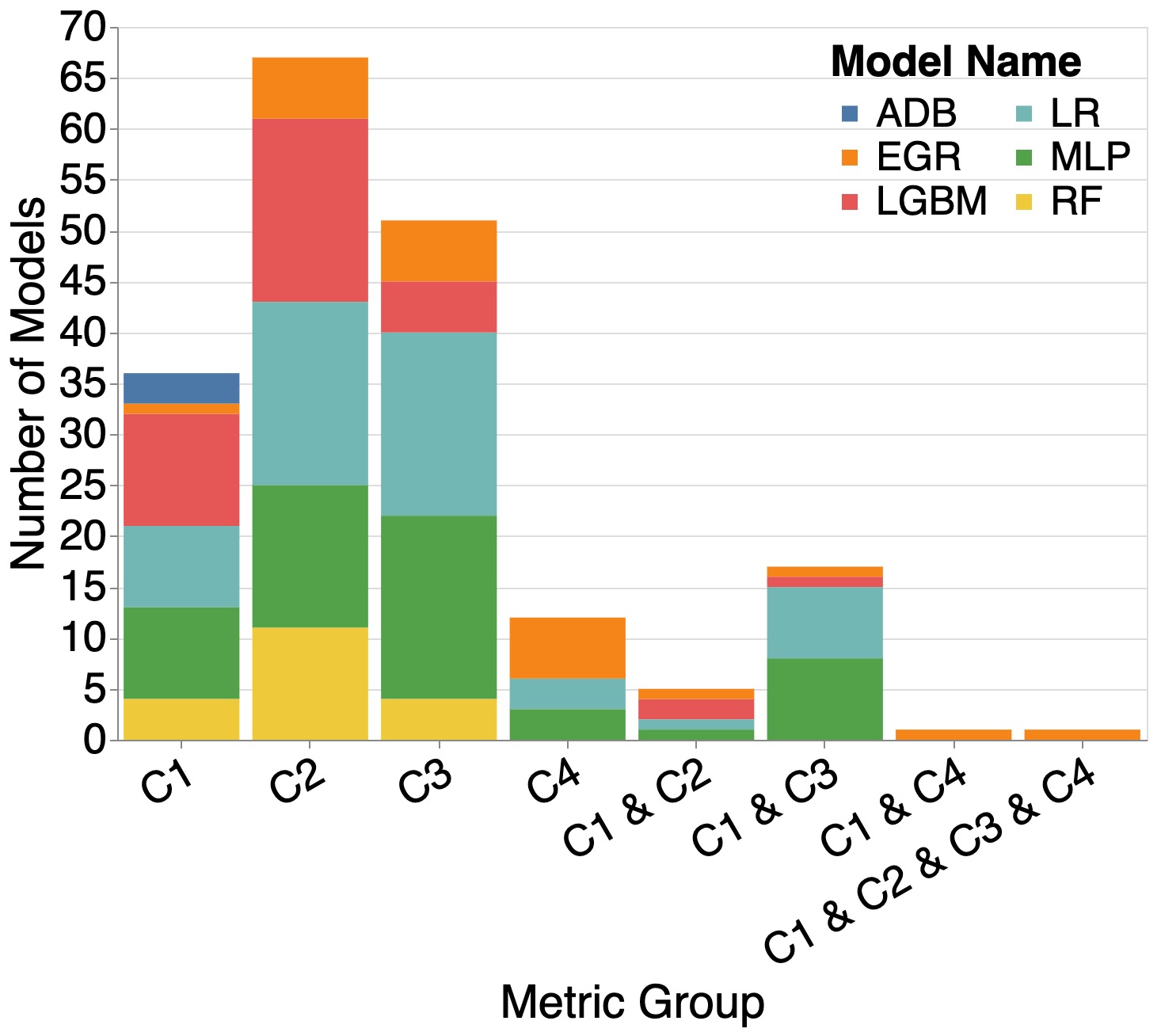}
         \caption{Model selection: There are several models with the desired accuracy (C1) and error-based fairness (C2), but only one with stability-parity (C3, C4)}
         \label{fig:model-selection-law}
     \end{subfigure}
     \hspace{10pt}
\caption{Law school. (a) Stability decompositions of different model types (b) Model selection with epistemic and aleatoric stability parity criteria. C1: Accuracy $\in$ [0.9, 1], C2: False Negative Rate Difference $\in$ [-0.02, 0.02], C3: Epistemic Uncertainty Difference $\in$ [-0.02, 0.02], C4: Aleatoric Uncertainty Difference $\in$ [-0.02, 0.02]. dt\_clf is Decision Tree, lr\_clf is Logistic Regression, rf\_clf is Random Forest, mlp\_clf is Neural Network, historically called the Multi-Layer Perceptron, lgbm\_clf is Gradient Boosted Trees, gandalf\_clf is a deep table-learning method. Adversarial Debiasing (ADB) and Exponentiated Gradient Reduction (EGR) are in-processing interventions and thereby constitute their own model type. }
\label{fig:law-school-intro}
\end{figure*}

Instability/arbitrariness has so far been studied in terms of disagreement in predicted labels between models in the Rashomon set~\cite{marx2020predictive, cooper2023arbitrariness, d2022underspecification, black_multiplicity_22}. In this work, we show that label instability can arise due to different reasons, namely from inconsistency or under-confidence. As a motivating example consider the task of credit allocation (binary yes/no) for three individuals $x_1$, $x_2$, and $x_3$ in Table ~\ref{tab:example}. $m_1$, $m_2$, and $m_3$ are models with equivalent accuracy. Assuming the usual cut-off of 0.5 for positive prediction, the prediction for $m_1$ is perfectly stable (3/3 models agree) whereas for $x_2$ and $x_3$ 2/3 models agree on their prediction. Existing multiplicity metrics such as self-consistency, label stability and ambiguity are defined in terms of label disagreement, and will treat $x_2$ and $x_3$ equivalently. However, notice that for $x_2$ the instability is because there is high variance in the predicted probability, whereas for $x_3$ it is because all of the predicted probabilities lie close to the decision boundary and so a small amount of variance will also flip the predicted label. Put differently, for $x_2$ instability arises from inconsistent predictions whereas for $x_3$ instability arises from under-confident (or uninformative) predictions.

\citet{coston2023validity} translate concepts from validity theory to examine when it is appropriate to use predictive systems in critical social contexts. Instability undermines validity because, when different models of comparable quality produce conflicting predictions—for instance, one recommending loan denial and another approval—it becomes unclear which model’s decision should be trusted or acted upon.
 Further, \citet{coston2023validity} distinguish harms due to invalidity from harms due to value misalignment. Let us revisit the example in Table \ref{tab:example}. What if $x_1$, $x_2$, $x_3$ are canonical examples of three different social groups, say men, White women and Black women? Error-based fairness metrics would capture neither gender nor disability-based disparity in instability, since the accuracy is the same in all three cases. \citet{creel2022algorithmic} have argued that arbitrariness is an ethical concern when it is systematic or disparate, and \citet{cooper2023arbitrariness, gomez2024algorithmic} have reported that arbitrariness does track protected group membership in practice. Existing arbitrariness metrics at the label level would flag discrimination against women since the disagreement among the predicted labels of the models is systematically higher for $x_1$ than $x_2$ and $x_3$. They would, however, overlook that the model is systematically more inconsistent for White women than other groups, and systematically less confident in its prediction for Black women than other groups. This motivates more fine-grained analysis of arbitrariness, which we undertake in this work.

Drawing from the Bayesian uncertainty framework~\cite{abdar2021review, hullermeier2021aleatoric}, we empirically show that stability/arbitrariness decomposes into epistemic and aleatoric components, capturing the consistency and confidence in predictions, respectively. The observation we make is that although epistemic and aleatoric uncertainty metrics have primarily been studied within a Bayesian setting~\cite{tahir_aleatoric_Fairness, lakshminarayanan2017simple, gal2022bayesian}, they are more generally defined with respect to the distribution over model parameters --- whether this distribution is learned in a model-centric manner (e.g., via Bayesian networks) or in a data-centric manner (e.g., through bootstrapping over the training set). With this in mind, for the setting of multiplicity, we measure epistemic uncertainty (EU) by the predictive variance over models trained on different subsets of the same data, and aleatoric uncertainty (AU) by the mean entropy over all classes (2 here, since this is binary classification). The entropy-maximizing distribution is the uniform distribution, and so models that are unable to distinguish between classes (uniform probability assigned to all) will have an aleatoric uncertainty equal to 1. For epistemic uncertainty, the intuition is that more complex model classes are more likely to be sensitive to the inclusion or exclusion of a single training data point, thereby resulting in more inconsistent predictions. 

In summary, the aleatoric component captures when instability is due to under-confident predictions ($x_3$), whereas the epistemic component captures when instability is due to inconsistent predictions ($x_2$). This decomposition is shown on the \law dataset in Figure~\ref{fig:law-metrics}, and is significant because (a) it connects Bayesian uncertainty with multiplicity/stability, enabling cross-setting insights, and (b) it has practical utility in that it helps constrain the set of ``good models'' by providing a more precise criterion, shown for the \law dataset in Figure~\ref{fig:model-selection-law}. 


\textbf{Summary of contributions} We built and publicly release a python library~\cite{herasymuk2024responsible} to measure epistemic and aleatoric instability in any ML pipeline, alongside existing confusion-matrix-based metrics, on the full test set and in terms of group-specific disparities (fairness metrics). 

\textbf{Our first contribution} is an extensive empirical evaluation of epistemic and aleatoric instability in popular fairness research datasets using state-of-the-art model architectures, yielding the following \textbf{insights}:
\begin{enumerate}
    \item Different model types have different epistemic and aleatoric decompositions of arbitrariness/instability, and each of these can be systematically higher for some demographic groups than others.
    \item State-of-the-art models have high aleatoric instability (58-79\%) and significant disparities in aleatoric instability (25-29\%) on popular fairness benchmarks. Epistemic instability is also fairly high (0.1-0.14) but is close to uniform for different groups (maximum disparity of 0.05, with perfect parity on 2/7 datasets). 
    \item Arbitrariness/instability behaves like uncertainty out of domain: epistemic instability is high on unseen data; and while epistemic instability can be reduced with more training data, aleatoric instability cannot.
    \item Fairness in-processing generally decreases aleatoric instability while fairness pre-processing generally increases it compared to the baseline (no intervention) for all model types. Epistemic instability is sensitive to both model type and intervention. Fairness interventions do not generally increase stability disparities beyond that of the baseline (no intervention).
\end{enumerate}

\textbf{Our second contribution} is a practical solution to the problem of systematic arbitrariness---a model selection procedure that includes epistemic and aleatoric stability criteria alongside existing accuracy and fairness criteria.  We show that this procedure successfully narrows down a large set of reasonable models (50-100 on our experimental tasks) to a handful of models that are stable, fair, and accurate.

\section{Background and Related Work}
\label{sec:related}

In their seminal paper, \citet{breiman_two_cultures} uncovered the \emph{Rashomon effect}: the phenomenon where a large number of models, called the \emph{Rashomon set}, can explain a given dataset equally well in terms of empirical loss. This is benign if these models produce similar predictions—but a large body of work shows they often do not~\cite{black_leave_one_out_Fairness, cooper2023arbitrariness, marx2020predictive, black_multiplicity_22, Watson_Daniels_Parkes_Ustun_2023}. \citet{paes2023inevitability} formally show how prevalent the Rashomon effect is for fixed datasets, especially when common data transformations amplify variation caused by initialization. \citet{watson2023predictive} explain that predictive multiplicity is greater for outliers, near-boundary examples, less separable datasets, and minority groups in imbalanced datasets. Common multiplicity metrics include ambiguity (the proportion of inputs with conflicting predictions)~\cite{marx2020predictive}, discrepancy (the maximum number of prediction changes across models)~\cite{marx2020predictive}, and pairwise disagreement (the fraction of disagreeing model pairs per input)~\cite{d2022underspecification, black_multiplicity_22}. \citet{hsu2022rashomon} define Rashomon Capacity using KL-divergence between model output distributions. 

\textbf{A multiplicity of multiplicities} This phenomenon has also been framed as \emph{underspecification} in ML pipelines~\cite{d2022underspecification}. \citet{meyer2023dataset} define \emph{dataset multiplicity} as the space of all plausible dataset variants, studying its effect on linear models with label noise. \citet{long2023arbitrariness} and \citet{cooper2023arbitrariness} assess fairness–arbitrariness trade-offs, showing that arbitrariness can remain high even under fairness constraints. \citet{long2023arbitrariness} further show that standard fairness metrics and interventions do not reduce multiplicity, while \citet{cooper2023arbitrariness} show that reducing arbitrariness via ensembling can improve error parity without explicit fairness interventions. \citet{cavus2024investigating} examine how rebalancing and variable selection impact stability. \citet{simson2024one} propose multiverse analysis to prevent “fairness hacking” in ML pipelines. \citet{chen2018my} decompose error-based discrimination into bias, variance, and noise, and propose methods to estimate and mitigate each. By contrast, we decompose instability into epistemic and aleatoric components. \citet{wang2024aleatoric} introduce notions of \emph{epistemic} and \emph{aleatoric} discrimination ---due to model design and inherent data bias, respectively --- but measure these via errors, not uncertainty itself. Unlike their fixed-data setting, we show that fairness pre-processing alters epistemic and aleatoric uncertainty. Moreover, while they focus on fairness–accuracy trade-offs, we emphasize stability and stability-parity.

We defer to \citet{ganesh2025curious} for a comprehensive review of multiplicity. To our knowledge, no prior work has explored the epistemic/aleatoric decomposition of instability/arbitrariness.

\textbf{Uncertainty} has been extensively studied in Bayesian neural networks~\cite{depeweg2018decomposition, pmlr-v48-gal16, nado2021uncertainty}, with applications in vision~\cite{gal_vision_neurips, Mukhoti_2023_CVPR, mukhoti2021deterministic}, language~\cite{xiao2019wat}, and physics~\cite{gal2022bayesian}. \citet{hullermeier2021aleatoric, der2009aleatory, malinin2019uncertainty} distinguish epistemic from aleatoric uncertainty. Ensembling for uncertainty quantification and reduction is well-explored~\cite{NEURIPS2021_rahaman_UQ, zhang2020mix}. See \cite{deep_UQ_review} for a broad survey of methods.

\citet{gummadi_uncertainty} and \citet{tahir_aleatoric_Fairness} link uncertainty and fairness. \citet{gummadi_uncertainty} first connected multiplicity to uncertainty, focusing on epistemic errors in ambiguous regions and improving error parity therein. \citet{tahir_aleatoric_Fairness} focus on aleatoric uncertainty, showing that low-uncertainty samples are more accurately modeled, and propose fairness interventions targeting high-uncertainty regions. While our definitions share Bayesian foundations~\cite{abdar2021review, hullermeier2021aleatoric, tahir_aleatoric_Fairness, lakshminarayanan2017simple}, their methods rely on Bayesian neural networks, whereas we use bootstrapped ensembling to directly capture predictive multiplicity.

\begin{figure*}[t]
    \centering
    \includegraphics[width=0.9\linewidth]{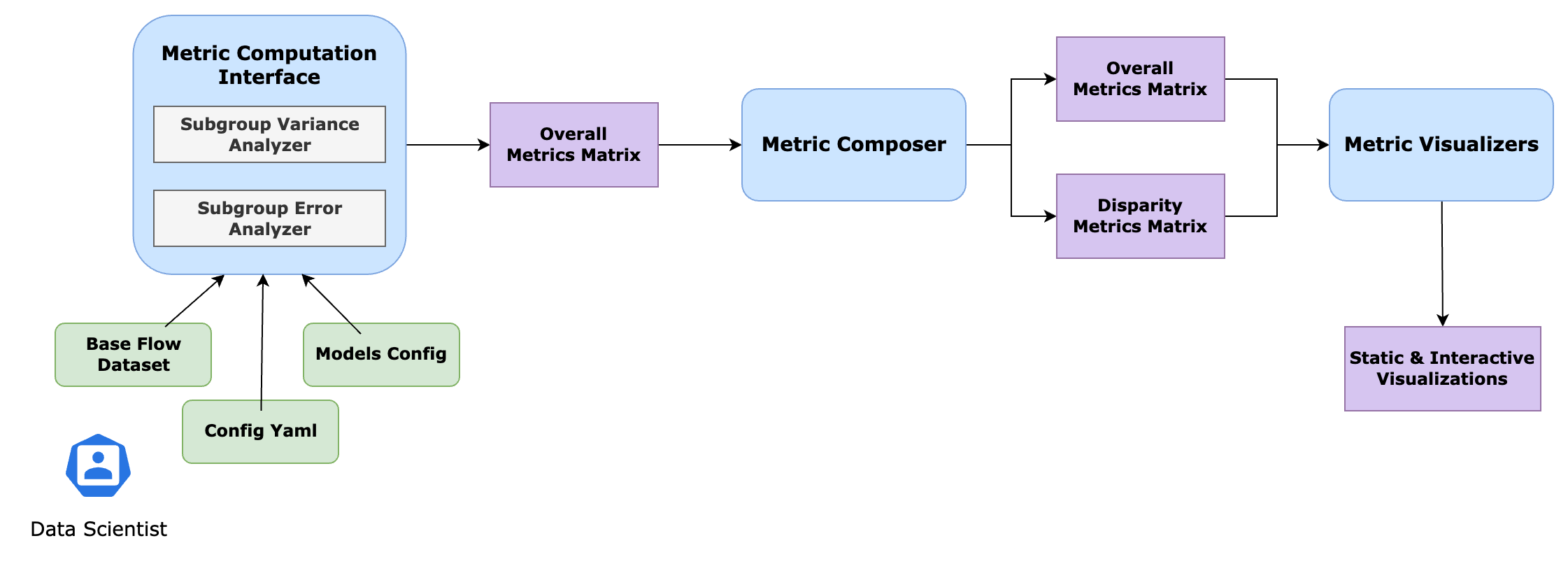}
    \caption{\libname Architecture: Inputs are shown in green, pipeline stages are shown in blue, and the output of each stage is shown in purple.}
    \label{fig:library_diagram}
\end{figure*}

\textbf{Harms of systematic arbitrariness} \citet{black_leave_one_out_Fairness} show that the inclusion/exclusion of a single data points can influence model predictions, framing such arbitrariness as unfair. \citet{creel2022algorithmic}, however, argue that only systematic (or disparate) arbitrariness is morally problematic. We adopt this view: fairness requires that predictions not be systematically more stable for one group than another. \citet{gomez2024algorithmic} explore harms from systematic arbitrariness in content moderation, identifying threats to freedom of expression, non-discrimination, and procedural justice. They argue that such systems violate due process, impartiality, and accountability. Model selection becomes a hidden lottery, exposing a subset of users to random treatment.

Ensembling remains the main strategy to reduce arbitrariness~\cite{zhang2020mix, NEURIPS2021_rahaman_UQ, long2023arbitrariness, individual_arbitrariness, cooper2023arbitrariness, black2022selective}. \citet{long2023arbitrariness, individual_arbitrariness} construct convex combinations; \citet{black2022selective} abstain when model consensus is low; \citet{cooper2023arbitrariness} adapt bagging with abstention. Notably, however, \emph{abstaining  does not solve the problem of arbitrariness, it merely shifts it}. For example, the ensemble may exhibit systematically higher abstention rates for one group, reflecting greater underlying arbitrariness for that group.
We propose incorporating stability criteria into model selection. This aligns with \citet{black2022less_discriminatory}, who advocate searching for less discriminatory models when accuracy is comparable, and with the “meta-rules” proposed by \citet{black_multiplicity_22}.

\section{Preliminaries}
\label{sec:prelims}

Consider a data-generating process $P$ from which we can sample a dataset $D_k=\{(x_1,y_1),…,(x_n,y_n)\}$, where $x_i \in X$ are covariates/features and $y_i \in Y$ are the observed labels. Further, given a hypothesis class $H$, let $h_{D_k}$ denote the model learned from dataset $D_k$, by some training procedure $f$. Each trained model $h_{D_k}: X \rightarrow Y $ maps a feature vector $x_i$ to a predicted label $y_i$. We restrict our focus to binary classification in this work, so assume $Y = \{0,1\}$\footnote{Our setup can be easily extended to multi-class prediction}. For notational simplicity, let's assume $h_{D_k}$ returns the probability of the positive class, instead of the predicted label. Labels can be determined by applying a suitable threshold to this predicted probability $\hat{y_i} = \textbf{1}[h_{D_k}(x_i)>=t]$, where $t=0.5$ usually. Note that $h_{D_k}$ produces a deterministic mapping, and so the sources of randomness in this set up are the training dataset $D_k$ and the training procedure $f$. Let's assume we sample $m$ datasets $D_1, D_2, ... D_m$ and fit $m$ models $h_{D_1}, h_{D_2}, ... h_{D_m}$. Now, we define stability metrics on a per-sample basis. We use $m$=200 in all of our experiments, and generate $D_1 ... D_m$ using the bootstrap~\cite{efron1994bootstrap}, where each $D_k$ is a random 80\% of the full training dataset. 

\subsection{Metrics}

\textbf{Label Stability}, closely related to the self-consistency metric from ~\citet{cooper2023arbitrariness} and the ambiguity metric from ~\citet{marx2020predictive}, is defined as the normalized absolute difference between the number of times a sample is classified as positive vs. negative~\cite{darling2018toward, khan2023fairness}:

\begin{equation}
\label{eqn:label-stability}
    LS_h(x) = \frac{1}{m} |\sum_{j=1}^m \mathbf{1}[h_{D_j}(x)>=t] - \sum_{i=1}^m \mathbf{1}[h_{D_i}(x)<t]|
\end{equation}

\textbf{Epistemic uncertainty} is quantified as the predictive standard deviation over all models~\cite{hullermeier2021aleatoric, tahir_aleatoric_Fairness}:
\begin{equation}
\label{eqn:epistemic}
     EU_h(x)^2 = \text{Var}_j[h_{D_j}(x)]
\end{equation}

\textbf{Aleatoric uncertainty} is quantified as the mean entropy, over all models~\cite{hullermeier2021aleatoric, tahir_aleatoric_Fairness}: 
\begin{equation}
\label{eqn:aleatoric}
    AU_h(x) = \E_j[h_{D_j}(x)log(h_{D_j}(x)) + (1-h_{D_j}(x))log(1-h_{D_j}(x))]
\end{equation}

\subsection{The \libname software library}
\label{sec:library}
To streamline the computation of stability and uncertainty metrics alongside traditionally reported confusion-matrix-based metrics, we developed a Python library for in-depth profiling of model performance across overall and disparity dimensions. The design of the library was steered by three core principles: 1) facilitating easy extensibility of model analysis capabilities; 2) ensuring compatibility with user-defined datasets and model types; 3)~enabling simple composition of parity metrics based on the context of use. The software library decouples the process of model profiling into several stages, which include \textit{subgroup metric computation}, \textit{disparity metric composition}, and \textit{metric visualization}. This separation provides data scientists with enhanced control and flexibility when utilizing the library, both during model development and for post-deployment monitoring. Figure~\ref{fig:library_diagram} illustrates how the library constructs a pipeline for model analysis. Inputs to a user interface are depicted in green, pipeline stages are represented in blue, and the output of each stage is shown in purple. Additional details about our software library~\cite{herasymuk2024responsible} are deferred to the Appendix.

\subsection{Tasks, Datasets and Protected Groups}
We use ACSIncome and ACSPublicCoverage from Folktables~\cite{folktables_ding2021}, Law-School~\cite{wightman1998lsac} and Student Performance~\cite{cortez2008using} as our primary datasets, with supplementary evaluation on  German\footnote{\url{https://archive.ics.uci.edu/dataset/144/statlog+german+credit+data}}~\cite{misc_statlog_(german_credit_data)_144}, Bank Marketing\footnote{\url{https://archive.ics.uci.edu/dataset/222/bank+marketing}}
~\cite{Moro2014ADA} and Diabetes\footnote{\url{https://www.kaggle.com/datasets/tigganeha4/diabetes-dataset-2019}}~\cite{diabetes_dataset} datasets. For each dataset, there is a binary classification task, where being assigned a positive label corresponds to the attainment of a desirable social good (such as education, employment, or access to healthcare). We selected these datasets from ~\citet{le2022survey} to have good coverage of (i) social domains and (ii) dataset sizes and other characteristics such as number of numerical and categorical features. We provide descriptions and summary statistics (base rates and proportions of different groups) of each dataset in the Appendix. 

\subsection{Model Types}
We evaluate 6 model types: decision tree (\texttt{dt$\_$clf}), logistic regression (\texttt{lr$\_$clf}), gradient boosted trees (\texttt{lgbm$\_$clf)}, random forest (\texttt{rf$\_$clf}), neural network, historically called the multi-layer perceptron (\texttt{mlp$\_$clf}), and a deep table-learning method called GANDALF~\cite{joseph2022gandalf} (\texttt{gandalf$\_$clf)}. Additional details about hyperparameter tuning for each model type are deferred to the Appendix. 
\section{Stability Decomposition}
\label{sec:criteria}

\begin{table*}[h]
\small 
\caption{Performance baselines (no interventions), averaged over all model types and 6 experimental runs, using m=200 models. $F1$, $LS$, $AU$, $EU$ are overall metrics, and the remaining are disparity (fairness) metrics. We report both the mean and the Std for the F1 to indicate the degree of multiplicity across model classes, with lower Std indicating more multiplicity.}
\begin{tabular}{ccccccccccccc}
 &   $F1$ &  $LS$ &  $AU$ &  $EU$ &  $\Delta Accuracy$ &  $PPVD$ &  $TPRD$ &  $TNRD$ &  $SRD$ & $\Delta LS$ &  $\Delta AU$ &  $\Delta EU$ \\
folk-income       & 0.70  $\pm$  0.02 &             0.87 &                   \textbf{0.58} & 0.07 &           0.04 &     -0.11 &     -0.03 &      0.02 &                -0.10 &                  0.02 &                       -0.05 &     -0.01 \\

law school & 0.95 $\pm$ 0.00 &             0.95 &                   0.32 & 0.04 &          -0.13 &     -0.11 &     -0.09 &      0.30 &                -0.18 &                 -0.09 &                        \textbf{0.29} &      0.03 \\

folk-pubcov & 0.53 $\pm$  0.02 &  0.85 &  \textbf{0.79} &  0.07 & -0.01 & -0.01 & -0.06 &  0.01 &
       -0.03 & -0.01 &  0.02 & \textbf{0.00} \\ 

student & 0.96 $\pm$  0.01 &  0.93 &  0.21 &  0.06 &  0.05 &  0.06 &  0.00  &  0.23 &
        0.02 &  0.03 & -0.06 & -0.02 \\

german     & 0.84 $\pm$ 0.02 &             0.74 &                   \textbf{0.60} & \textbf{0.14} &          -0.14 &     -0.20 &     -0.06 &      0.01 &                -0.10 &                 -0.07 &                        0.05 &      0.01 \\

bank       & 0.26 $\pm$    0.07 &             0.94 &                   0.39 & 0.05 &          \textbf{-0.23} &     -0.01 &      0.12 &     -0.11 &                 0.15 &                 \textbf{-0.17} &                        \textbf{0.25} &      0.05 \\

diabetes   & 0.85 $\pm$    0.06 &             0.88 &                   0.32 & \textbf{0.10} &           0.04 &      0.07 &      0.15 &      0.01 &                 0.10 &                \textbf{-0.00} &                        \textbf{0.00} &     \textbf{-0.00} \\
\end{tabular}
\vspace{0.2cm}
\label{tab:performance_baselines}
\end{table*}

\begin{figure}[h]
\begin{subfigure}[h]{\linewidth}
    \centering
    \includegraphics[width=\linewidth]{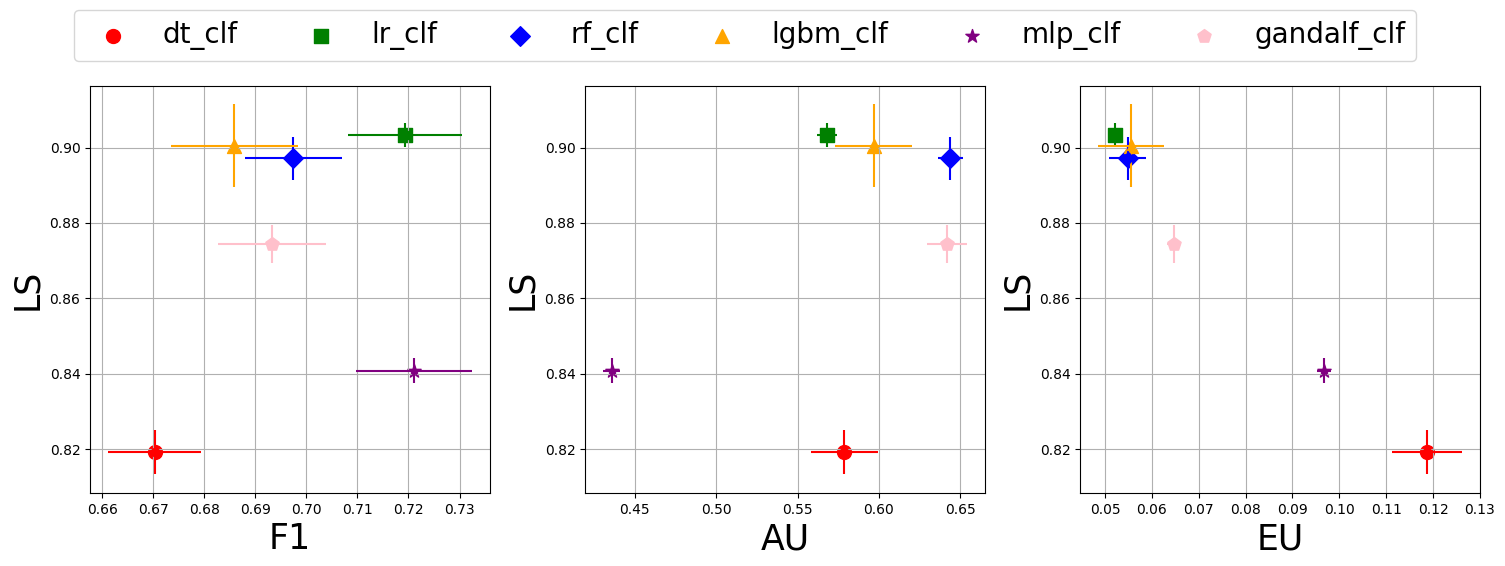}
    \caption{Folk-Income}
\end{subfigure}
\begin{subfigure}[h]{\linewidth}
    \centering
    \includegraphics[width=\linewidth]{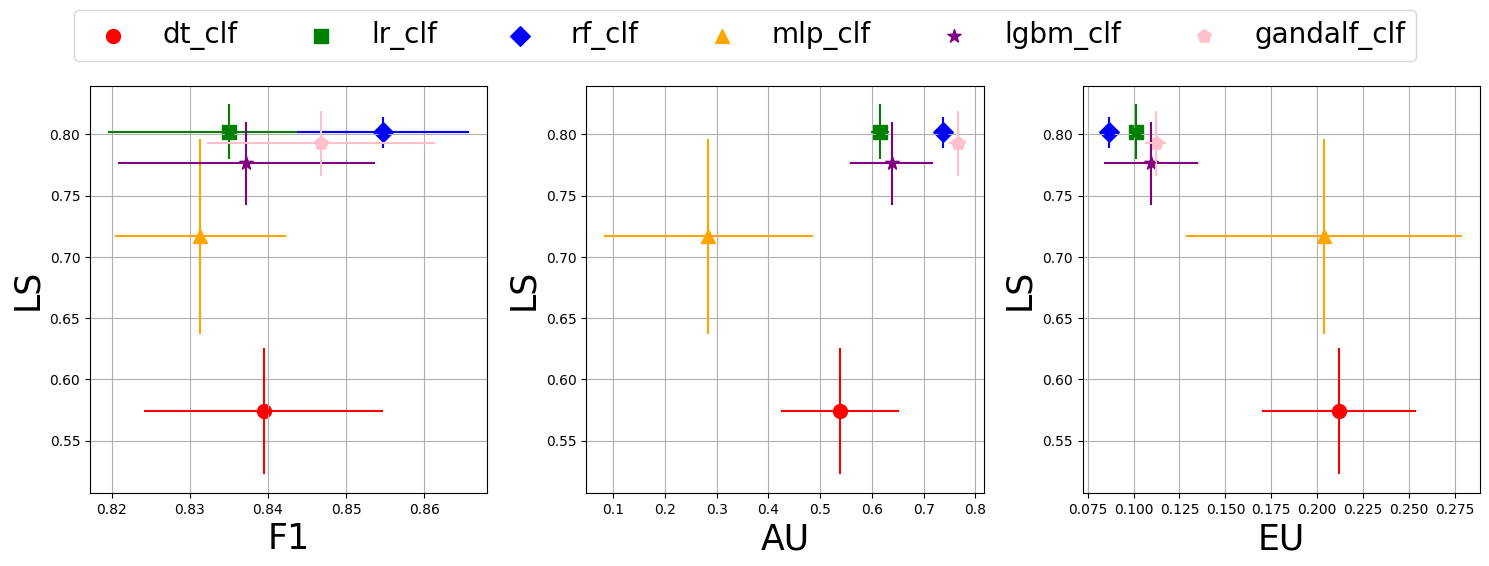}
    \caption{German}
\end{subfigure}
\caption{Stability decomposition: Label stability is affected by both epistemic uncertainty and aleatoric uncertainty.}
\label{fig:stability-decomposition}
\end{figure}

We demonstrate the empirical decomposition of stability into epistemic and aleatoric components for different model types (markers) on Folk-income and German datasets in Figure~\ref{fig:stability-decomposition}, with results on other datasets deferred to Figure~\ref{fig:stability-tradeoffs} in the Appendix. 
Different model architectures admit different epistemic and aleatoric stability decompositions. All models exhibit similar F1 scores and low aleatoric uncertainty (except for the Decision Tree), indicating that differences in label stability are driven by epistemic uncertainty. In contrast, on the German dataset, aleatoric uncertainty is so high that it overrides epistemic effects and dominates label stability across all model types.

We find that logistic regression has the lowest epistemic uncertainty (Std) on all datasets, corroborating our earlier discussion about simpler models being more consistent. Ensemble models (RF and LGBM) are generally more consistent than decision trees. It is unclear, however, how to compare the complexity of these model types with neural archictectures. For example, surprisingly, GANDALF has lower epistemic uncertainty than DT on all datasets. 

GANDALF generally exhibits the highest aleatoric uncertainty across all datasets, but no consistent trend is observed across other model types.

\begin{figure*}[h!]
    \centering
    \includegraphics[width=\linewidth]{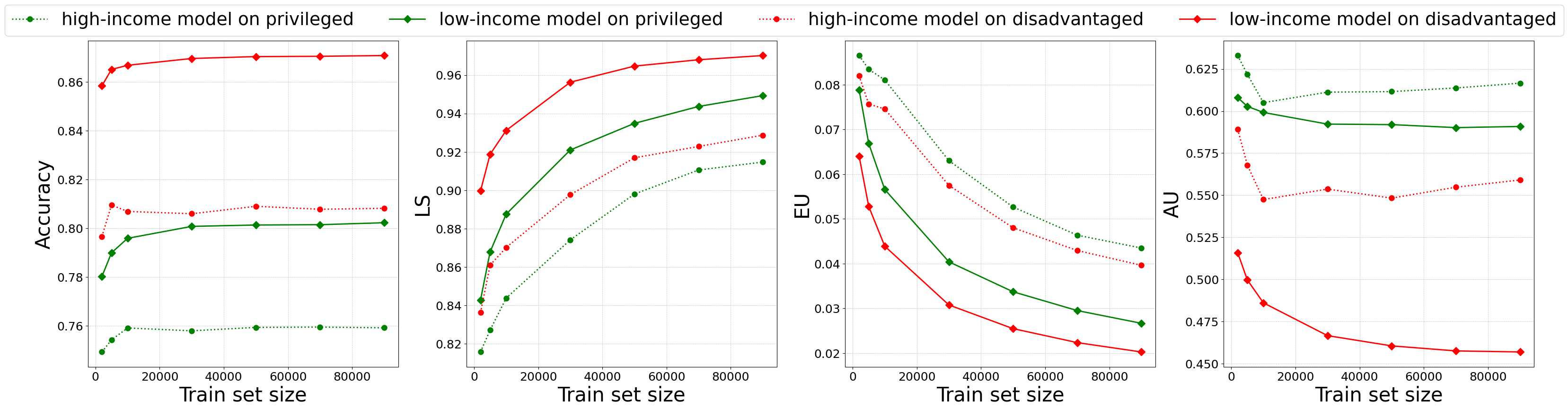}
    \caption{Testing on the low income domain. In-domain models (trained on low income domain) are showed in bold lines, out-domain models (trained on the high income domain) are shown in dashed lines. Average performance on the privileged group is shown in green, and on the disadvantaged in red.}
    \label{fig:low-income-testing}
\end{figure*}

\subsection{Stability trade-offs in real-world datasets}
Next, we look at how stability trades off with other desiderata. In Table~\ref{tab:performance_baselines} we report (i) the F1 score; (ii) stability metrics: Label Stability (LS), Aleatoric Uncertainty (AU), and Epistemic Uncertainty (EU); (iii) classic fairness metrics: Accuracy Difference ($\Delta Accuracy$)~\cite{corbett2023measure}, Positive Predictive Value Difference ($PPVD$, also called Calibration Difference)~\cite{chouldechova_impossibility, Kleinberg_impossibility}, True Positive Rate Difference ($TPRD$)~\cite{hardt2016equality}, True Negative Rate Difference ($TNRD$)~\cite{hardt2016equality}, and Selection Rate Difference ($SRD$)~\cite{feldman2015certifying}; and (iv) stability-based fairness metrics: Label Stability Difference ($\Delta LS$), Aleatoric Uncertainty Difference ($\Delta AU$), and Epistemic Uncertainty Difference ($\Delta EU$), defined analogously to existing group fairness metrics as the difference in the average value of the metric on disadvantaged and privileged subsets. We report the average of each metric over all 6 model types and 6 experimental runs, on different datasets in Table~\ref{tab:performance_baselines}. 

We find evidence of significant instability in real-world datasets, specifically: high aleatoric uncertainty on \folk (0.58), \folkcov (0.79) and \german (0.60), and high epistemic uncertainty on \german (0.14) and \diabetes (0.10). We only find evidence of systematic instability disparity on \law ($\Delta AU$=0.29) and \bank ($\Delta LS$=-0.17, $\Delta AU$=0.25). Notably, all models  show perfect stability-parity on \diabetes.

\section{Uncertainty Behavior Out-of-Distribution}

\textbf{Experiment Goals} The uncertainty framework distinguishes modeling uncertainty (`epistemic' derived from `knowledge') from data uncertainty (`aleatoric' derived from `dice' or chance). Epistemic uncertainty therefore is reducible, for example, by collecting more data, whereas aleatoric uncertainty is due to the inherent variability in the data and is thereby irreducible. We empirically evaluate whether epistemic and aleatoric instability also demonstrate these properties. 

\begin{figure*}[h]
    \centering
    \includegraphics[width=\linewidth]{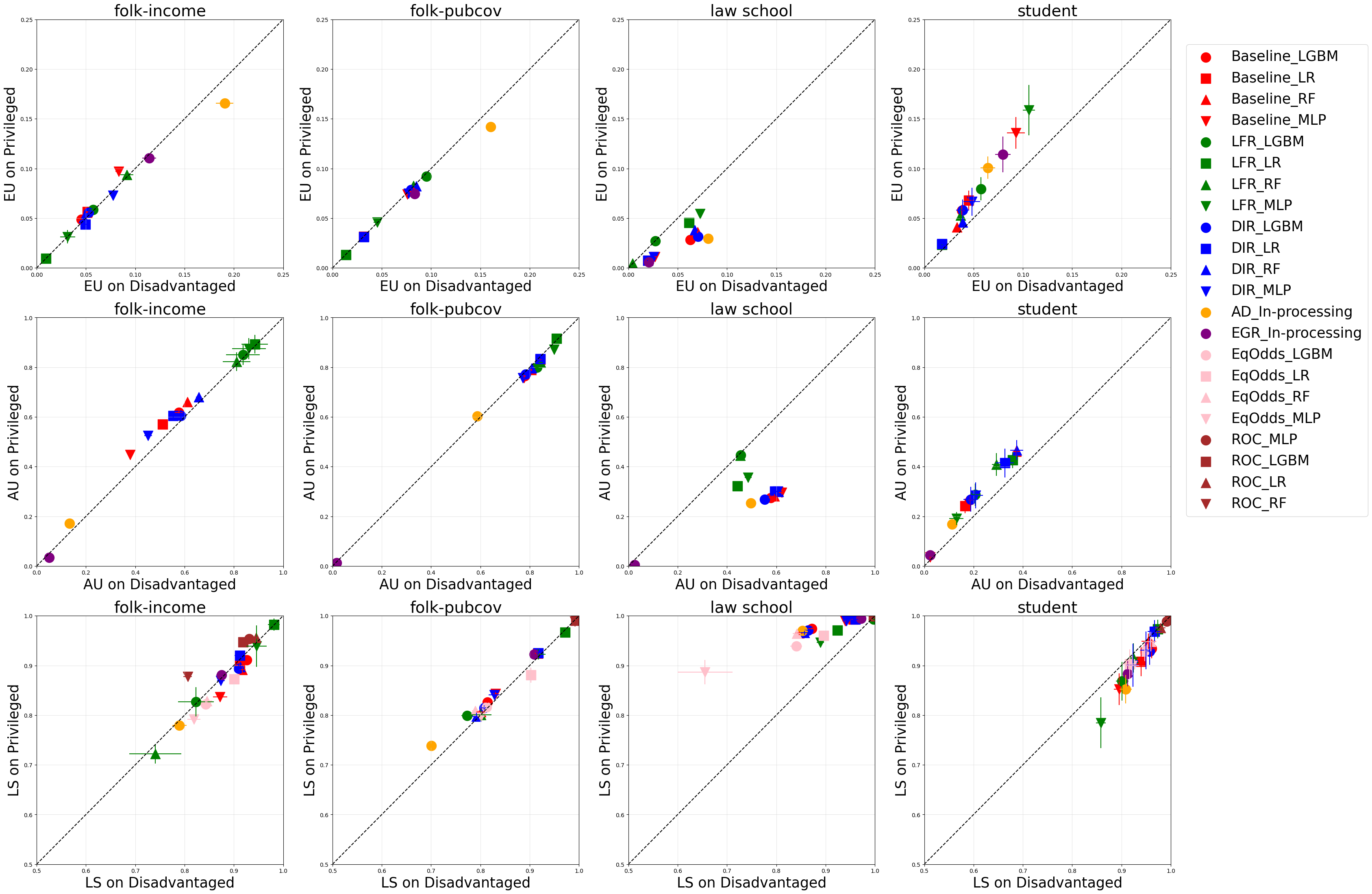}
    \caption{Fairness interventions: Columns are datasets, rows are metrics. Colors indicate interventions and shapes indicate model types. For each metric the average value on the disadvantaged group is reported on the x-axis and for the privileged group on the y-axis. For EU and AU lower values are better. For LS higher values are better. Moving along the y=x line indicates that interventions improve or worsen stability uniformly for all groups, whereas moving off the y=x line indicates a stability disparity. The baseline indicates no fairness intervention.}
    \label{fig:fairness-interventions}
\end{figure*}

\textbf{Setup} We construct an Out-of-Distribution (OOD) prediction task on Folktables~\cite{folktables_ding2021}, using the fact that it contains data from all 50 US states for each year. We use the ACSIncome task, and construct two data domains: (1) high-income domain: constructed by aggregating data from the states of MD, NJ, and MA and (2) low-income domain: from the states of WV, MS, AR, NM, LA, AL, and KY. To explore the impact of different training set sizes, we employ sets containing 2K, 5K, 10K, 30K, 50K, 70K, and 90K samples. We train LR models, once on data from the high-income domain and once on the low-income domain, and test them on both domains. Importantly, the in-domain and out-of-domain test sets have the same 10K size and include the same samples across experiments with different train set sizes. We tune each model using 3-fold cross-validation.

\textbf{Findings} We report performance on the low income test set in Figure~\ref{fig:low-income-testing}, and defer results on the high income test set to the Appendix. We find that out-of-domain models (shown with a dashed line) always have higher epistemic uncertainty than in-domain models (shown with a bold line). Increasing training set size improves model stability, but only by reducing epistemic uncertainty. Aleatoric uncertainty converges at about 50k samples, whereas epistemic uncertainty continues to decrease as the model is trained on larger quantities of data. These findings validate the expected behavior of the uncertainty metrics for the setting of predictive multiplicity: (i) models have high epistemic uncertainty in unseen data domains; (ii) epistemic uncertainty is reducible with more data, whereas aleatoric uncertainty is not. Interestingly, we find that models are more accurate and stable on the disadvantaged group: the model trained on the low-income domain has larger stability disparities, and increasing the training set size does not decrease the stability disparity.
\section{Effect of Fairness Interventions}
\label{sec:fairness interventions}

\textbf{Experiment Goals} Prior work has studied how different fairness interventions affect arbitrariness/multiplicity~\cite{cooper2023arbitrariness, long2023arbitrariness}. We are interested to study the effect on epistemic and aleatoric components. Fairness interventions are designed to optimize for existing criteria (error-parity, for example),  but this could come at the cost of increasing disparity in instability (by increasing disparity in the variance component of the error, for example). Intuitively, one would expect pre-processing interventions to affect aleatoric instability the most since they are both data-centric, and in-processing interventions to affect epistemic instability the most since they are both model-centric. Post-processing interventions only return predicted labels and not probabilities, so we only study label stability, and whether it becomes disparate, for post-processing interventions. 

\textbf{Setup} We evaluated 6 fairness interventions using IBM's AIF 360 toolkit~\cite{bellamy2019ai}, namely, Disparate Impact Remover (DIR)~\cite{feldman2015certifying} and Learning Fair Representations (LFR) ~\cite{zemel2013learning}, which are \textbf{pre-processing} interventions; Adversarial Debiasing (ADB) \cite{zhang2018mitigating} and Exponentiated Gradient Reduction (EGR) \cite{agarwal2018reductions}, which are \textbf{in-processing} interventions; and Equalized-Odds Postprocessing (EOP)~\cite{hardt_EOP2016} and Reject Option Classification (ROC) \cite{kamiran2012decision}, which are \textbf{post-processing} interventions. See Appendix for details. 

In Figure~\ref{fig:fairness-interventions} we report the average label stability, epistemic uncertainty and aleatoric uncertainty for the privileged (y-axis) and disadvantaged (x-axis) groups, for different model types (shapes) and fairness interventions (colors). The baseline (no fairness intervention) is in red. Moving along the y=x lines means the intervention transforms stability uniformly for privileged and disadvantaged groups, falling off this line indicates a stability disparity. We find that there is significant instability in all four datasets (with EU in the order 0.1).
 
\textbf{Findings} We do not see clusters by shape or by colors indicating that both model type and fairness intervention affect epistemic, aleatoric and label instability. Fairness in-processing generally decreases aleatoric uncertainty while fairness pre-processing generally increases it compared to the baseline (no intervention) for all model types. Epistemic instability is highly sensitive to both the intervention (color) and the model type (marker). 

In \folk and \folkcov, we observe close to perfect parity for all three stability metrics, for all interventions and model types. Fairness interventions on these datasets do not generally make the instability disparate — they increase or decrease it uniformly for both groups.  The instability, however, is disparate on the \law and \student datasets, with baseline models having higher epistemic uncertainty, higher aleatoric uncertainty and lower label stability for the disadvantaged group than privileged one, and fairness interventions such as EOP increase this stability-disparity, while LFR reduces it. Notably, the EGR in-processing technique is highly effective at mitigating aleatoric uncertainty (close to 0) on all datasets. This intervention also shows good stability-parity on all three metrics (with the exception of Std on \student). While there is no discernible trend about which fairness intervention is most/least stable, notably in all four datasets there is at least one intervention and model type combination that reaches perfect stability (and stability parity), indicating that by incorporating stability criteria during model development (evaluation and selection) we can `find' these fairly-stable models. We demonstrate this in the next section.
\begin{figure}[b]
    \centering
    \includegraphics[width=0.7\linewidth]{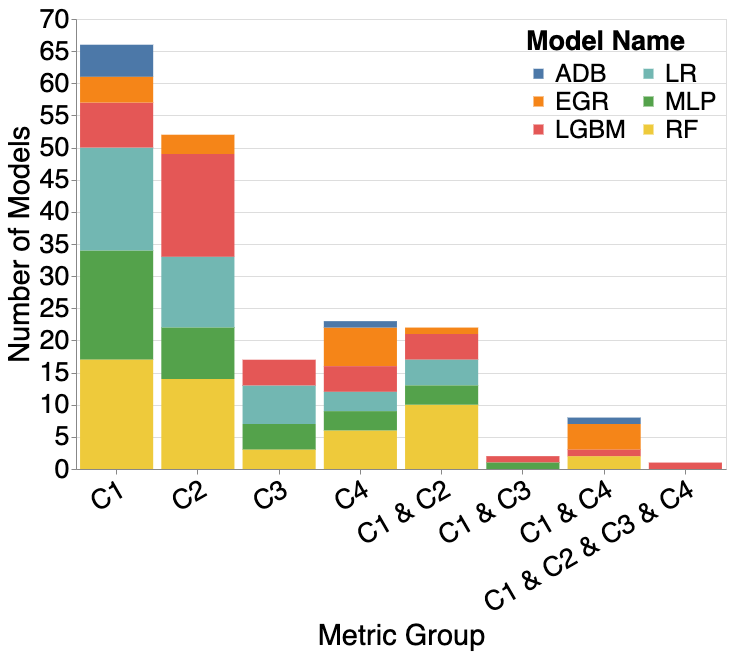}
    \caption{\folk: C1: Accuracy $\in$ [0.8, 1], C2: FPRD $\in$ [-0.02, 0.02], C3: $\Delta EU$ $\in$ [-0.001, 0.001], C4: $\Delta AU$ $\in$ [-0.02, 0.02]}
    \label{fig:model-selection-folk-income}
\end{figure}

\section{Model Selection with Stability Criteria}
\label{sec:model-selection}

Multiplicity has thus far been studied in terms of models that are indistinguishable in terms of accuracy. However, models can also be indistinguishable according to other desiderata, such as arbitrariness/stability, as we demonstrated empirically in real world datasets. \citet{black2022less_discriminatory} propose a broader reading of the Disparate Impact doctrine, compelling companies to look for less discriminatory alternatives (according to some suitable critiera), when there are several comparably accurate models. We demonstrate one instantiation of this idea by incorporating epistemic and aleatoric stability parity criteria (C3, C4) alongside existing error-based fairness criteria (C2) and accuracy (C1), shown for the models trained with fairness interventions on the folk-income task in Figure~\ref{fig:model-selection-folk-income}, with results for other datasets in Figure~\ref{fig:model-selection-full} in the Appendix. Note: adversarial debiasing (ADB) and exponentiated gradient reduction (EGR) are in-processing interventions, and thereby constitute their own model types. We apply each constraint (C1-4) first individually, then intersectionally. We can observe from Figure \ref{fig:model-selection-folk-income} that this procedure is effective at constraining a large set of accurate models to a few that are accurate, fair and stable.

\begin{figure}[h!]
     \centering
     \begin{subfigure}[h]{0.5\textwidth}
         \includegraphics[width=\textwidth]{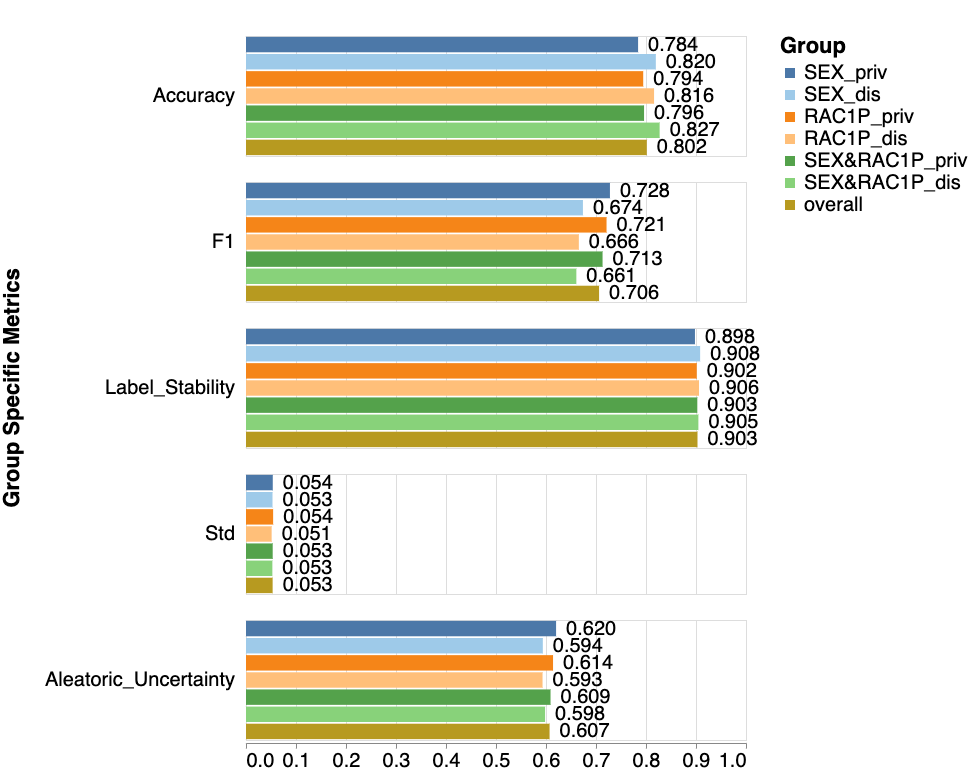}
     \end{subfigure}
     \hfill
     \begin{subfigure}[h]{0.5\textwidth}
         \includegraphics[width=\textwidth]{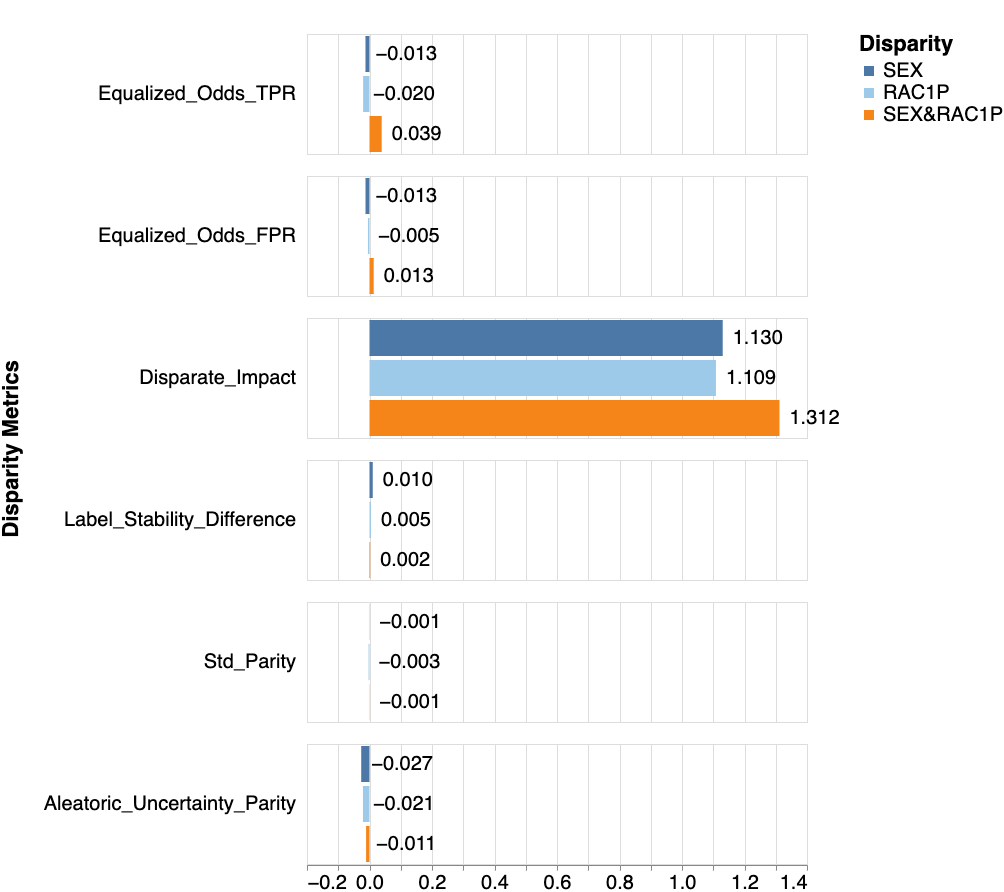}
     \end{subfigure}
\caption{Nutritional label for the LGBM selected on folk-income}
\label{fig:folk-income-nutritional-label}
\end{figure}

Importantly, fairness is a complex, non-technical concept that cannot be fully captured by mathematical criteria. While metrics can identify undesirable model behavior, it is unclear how to translate mathematical disparities into practical social harm. Which metrics to include during model selection and what acceptable ranges for them are is a context-specific question that falls under the mandate of relevant stakeholders and not ML researchers. Nonetheless, the path towards answering this question will necessarily go through the process of quantifying how much disparity we observe in practice. With this in mind, we provide a UI for interactive model selection in our \libname software library that allows practitioners to customize the constraints C1-4 from the extensive list of metrics the library computes and the ranges of disparities observed for their application. 

Furthermore, our software generates `nutritional labels' for the selected models, shown in Figure~\ref{fig:folk-income-nutritional-label} for the LGBM model selected on \folk in Figure~\ref{fig:model-selection-folk-income}. Recall from Figure~\ref{fig:model-selection-folk-income} we applied epistemic uncertainty parity and aleatoric uncertainty parity as secondary constraints (C3, C4) alongside accuracy (C1) and error-based fairness (C2). Unsurprisingly, the nutritional label of the selected model shows that the model has comparable stability on all demographic subgroups (by both single-axis and intersectional group definitions). As is commonly observed in practice, however, we see trade-offs with other fairness criteria, such as Disparate Impact (which is as high as 1.3 for intersectional subgroups). Such trade-offs underscore the need to evaluate and disclose model performance holistically using a large set of metrics, instead of relying on any single one entirely. 
\begin{figure*}[h!]
    \centering
    \includegraphics[width=0.8\linewidth]{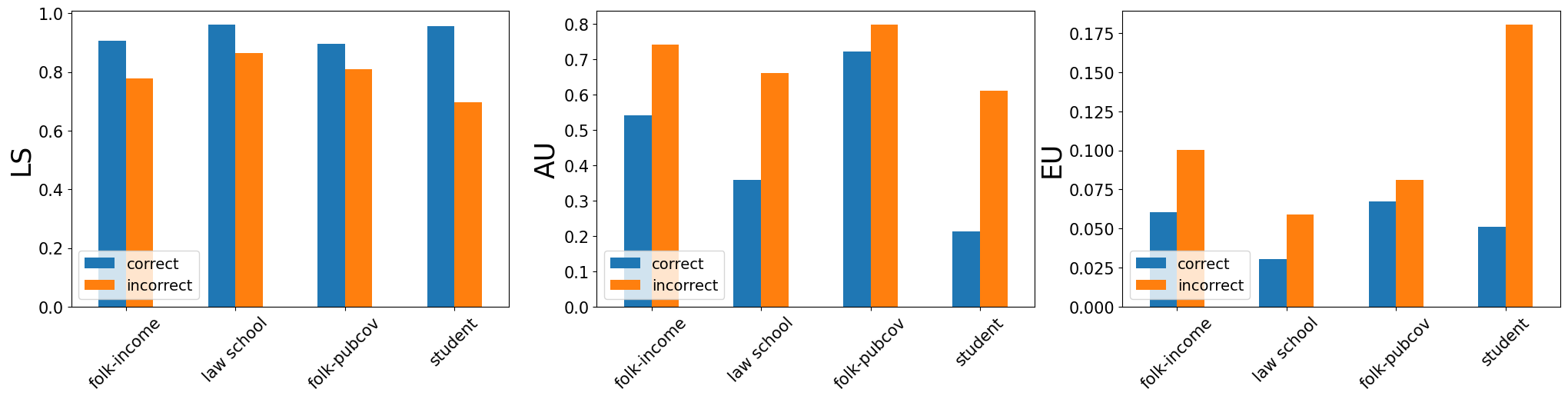}
    \caption{Models are more unstable on samples that they are wrong on. The average value of each metric, over all samples in the test set and all model types is reported. }
    \label{fig:case-analysis}
    \vspace{-0.5cm}
\end{figure*}

\section{Discussion}

\par{\textbf{From selecting models towards selecting pipelines}} Our empirical findings indicate that different model types admit different epistemic and aleatoric decompositions of instability, and popular fairness interventions transform this instability in unpredictable and unexpected ways. This indicates that we need to think about uncertainty quantification and mitigation not just during model training, but more holistically over all stages of the ML pipeline.  Our model selection procedures also need to become pipeline-aware, ie. we must move from selecting good models, towards selecting good pipelines, where stability is an additional dimension of responsibility.

\par{\textbf{Strengths and limitations of the Bayesian Uncertainty framework}} A common criticism of the uncertainty framework is its limited practical applicability~\cite{der2009aleatory, hullermeier2021aleatoric}. For example: in the context of ML models, the model is a function of the training data, and so it is impossible to cleanly isolate model-centric and data-centric uncertainty. In this work, however, we demonstrated a different strength of the uncertainty framework, ie. it's ability to distinguish between inconsistent and underconfident predictions. 

\par{\textbf{Models are more unstable on samples that they are wrong on}}
Purely mathematically, accuracy and stability are orthogonal: even a model with 99\% accuracy can have 100\% ambiguity, i.e., we can construct different models such that each makes that 1\% error on a different data point. In practice, we find that models are more unstable on samples that they make incorrect predictions for, shown in Figure~\ref{fig:case-analysis}. This indicates that propositions like ~\cite{tahir_aleatoric_Fairness} and ~\cite{gummadi_uncertainty}---which focus on reducing aleatoric and epistemic uncertainty on incorrectly classified samples---are a promising strategy for improving fairness according to error-parity. 

\par{\textbf{Social harm from systematic arbitrariness}} \citet{creel2022algorithmic} argue that arbitrariness by itself is not morally problematic, it is systematic arbitrariness that is of moral concern. \citet{gomez2024algorithmic} examine the social harms of systematic arbitrariness in the specific context of algorithmic content moderation, and find that it violates procedural justice and due process since it is impossible to know if a given prediction is due to multiplicity/arbitrariness, and this being higher for some groups than others violates non-discrimination doctrines by essentially creating a lottery on which data points (groups) will be subject to random treatment and which will not. Systematic arbitrariness has also been connected to other forms of outcome unfairness --- in explanations and recourse. Consider applicants' ability to seek recourse; If an applicant asks a decision-maker to explain why they received a certain outcome, then (1) the reliability of the explanation is affected by the stability of the rule --- how do we reliably explain an arbitrary decision-making procedure?~\cite{indeterminacy_explanations_zemel}
(2) the effectiveness of the recourse offered to applicants is affected by the stability-disparity of the rule: recourse will be less effective for members of the disadvantaged group because even if they improve their qualifications, the systematically higher arbitrariness of the rule towards them makes it uncertain whether they will, in fact, receive the desired outcome.~\cite{jiang2023recourse_multiplicity} 

We found significant arbitrariness/stability disparities in our empirical evaluation; most notably aleatoric instability disparities as high as 25\% and 29\% on the bank marketing and law school admission tasks, indicating that the risk of systematic arbitrariness from AI systems undermining due process in the distribution of critical social (financial and educational) goods is a practical reality, and not merely a hypothetical example. 

\par{\textbf{Transparency and disclosure mechanisms should include stability measurements}} Algorithmic transparency tools and other disclosure mechanisms seldom report model performance along dimensions such as stability and uncertainty. New legislation such as the EU AI Act does require stronger transparency commitments from AI vendors, but provides little guidance on what should be included in these transparency reports, and establishes no standards for what is considered `safe to deploy'. Tools such as model cards, nutritional labels, and data sheets have started to become the norm for the responsible release of new AI artifacts~\cite{gebru2021datasheets,holland2020dataset,mitchell2019model,stoyanovich2019nutritional}. However, the utility of these tools critically depends on the information disclosed in them, not simply in their existence. Model cards, for example, should also report stability and uncertainty metrics. We see our work as a necessary first step towards strengthening these transparency instruments. Only by reporting model performance along these dimensions will we be able to detect and document unfairness due to arbitrariness, building a case for legal protection against algorithmic instability.
\section{Conclusion}
\label{sec:conclusion}

Drawing from the uncertainty literature, we showed empirically that stability decomposes into epistemic and aleatoric components, capturing the consistency and confidence in prediction, respectively. Through an extensive empirical evaluation, we found that (i) state-of-the-art ML models have aleatoric instability as high as 79\% and aleatoric instability disparities as high as 29\% in popular fairness datasets; (ii) epistemic instability can be reduced with more training data whereas aleatoric instability cannot;  and (iii) both epistemic and aleatoric instability are highly sensitive to state-of-the-art fairness interventions. 

As a practical solution to this problem of systematic arbitrariness we demonstrated that incorporating secondary stability parity criteria during model selection alongside existing accuracy and error-based fairness criteria effectively reduces a large set of ``good'' models to a handful of accurate, stable and fair ones. 

 \paragraph{Limitations and Future Work} Our empirical demonstration of the epistemic and aleatoric decomposition of arbitrariness makes a connection between the Bayesian uncertainty literature and arbitrariness/multiplicity literature, which presents the opportunity to share insights between settings, for designing mitigation techniques, for example. We used the bootstrap procedure in our experiments because it produces reliable estimates from the context of model-agnostic sampling-based uncertainty quantification and is also a commonly used approach to simulating multiplicity, which allows for cross-setting insights, but this can be a computationally expensive procedure for large datasets and complex model architectures. It is exciting future work to speed up our procedure to compute stability and uncertainty metrics more efficiently, and to extend our empirical evaluation to other procedures for simulating multiplicity (random seeds, feature selection, etc). 
 
 Our epistemic and aleatoric decomposition of arbitrariness also raises new and interesting normative questions about whether systematic (or disparate) underconfidence and inconsistency pose the same ethical risks as systematic arbitrariness, which can be explored in future work.

\bibliography{main}

\begin{thebibliography}{75}
\providecommand{\natexlab}[1]{#1}

\bibitem[{Abdar et~al.(2021{\natexlab{a}})Abdar, Pourpanah, Hussain, Rezazadegan, Liu, Ghavamzadeh, Fieguth, Cao, Khosravi, Acharya, Makarenkov, and Nahavandi}]{deep_UQ_review}
Abdar, M.; Pourpanah, F.; Hussain, S.; Rezazadegan, D.; Liu, L.; Ghavamzadeh, M.; Fieguth, P.; Cao, X.; Khosravi, A.; Acharya, U.~R.; Makarenkov, V.; and Nahavandi, S. 2021{\natexlab{a}}.
\newblock A review of uncertainty quantification in deep learning: Techniques, applications and challenges.
\newblock \emph{Information Fusion}, 76: 243--297.

\bibitem[{Abdar et~al.(2021{\natexlab{b}})Abdar, Pourpanah, Hussain, Rezazadegan, Liu, Ghavamzadeh, Fieguth, Cao, Khosravi, Acharya et~al.}]{abdar2021review}
Abdar, M.; Pourpanah, F.; Hussain, S.; Rezazadegan, D.; Liu, L.; Ghavamzadeh, M.; Fieguth, P.; Cao, X.; Khosravi, A.; Acharya, U.~R.; et~al. 2021{\natexlab{b}}.
\newblock A review of uncertainty quantification in deep learning: Techniques, applications and challenges.
\newblock \emph{Information Fusion}, 76: 243--297.

\bibitem[{Agarwal et~al.(2018)Agarwal, Beygelzimer, Dud{\'\i}k, Langford, and Wallach}]{agarwal2018reductions}
Agarwal, A.; Beygelzimer, A.; Dud{\'\i}k, M.; Langford, J.; and Wallach, H. 2018.
\newblock A reductions approach to fair classification.
\newblock In \emph{International conference on machine learning}, 60--69. PMLR.

\bibitem[{Ali, Lahoti, and Gummadi(2021)}]{gummadi_uncertainty}
Ali, J.; Lahoti, P.; and Gummadi, K.~P. 2021.
\newblock Accounting for Model Uncertainty in Algorithmic Discrimination.
\newblock In \emph{Proceedings of the 2021 AAAI/ACM Conference on AI, Ethics, and Society}, AIES '21, 336–345. New York, NY, USA: Association for Computing Machinery.
\newblock ISBN 9781450384735.

\bibitem[{Bellamy et~al.(2019)Bellamy, Dey, Hind, Hoffman, Houde, Kannan, Lohia, Martino, Mehta, Mojsilovi{\'c} et~al.}]{bellamy2019ai}
Bellamy, R.~K.; Dey, K.; Hind, M.; Hoffman, S.~C.; Houde, S.; Kannan, K.; Lohia, P.; Martino, J.; Mehta, S.; Mojsilovi{\'c}, A.; et~al. 2019.
\newblock AI Fairness 360: An extensible toolkit for detecting and mitigating algorithmic bias.
\newblock \emph{IBM Journal of Research and Development}, 63(4/5): 4--1.

\bibitem[{Black and Fredrikson(2021)}]{black_leave_one_out_Fairness}
Black, E.; and Fredrikson, M. 2021.
\newblock Leave-one-out Unfairness.
\newblock In \emph{Proceedings of the 2021 ACM Conference on Fairness, Accountability, and Transparency}, FAccT '21, 285–295. New York, NY, USA: Association for Computing Machinery.
\newblock ISBN 9781450383097.

\bibitem[{Black et~al.(2023)Black, Koepke, Kim, Barocas, and Hsu}]{black2022less_discriminatory}
Black, E.; Koepke, J.~L.; Kim, P.; Barocas, S.; and Hsu, M. 2023.
\newblock Less Discriminatory Algorithms.
\newblock In \emph{Washington University in St. Louis Legal Studies Research Paper Forthcoming}.

\bibitem[{Black, Leino, and Fredrikson(2022)}]{black2022selective}
Black, E.; Leino, K.; and Fredrikson, M. 2022.
\newblock Selective Ensembles for Consistent Predictions.
\newblock In \emph{International Conference on Learning Representations}.

\bibitem[{Black, Raghavan, and Barocas(2022)}]{black_multiplicity_22}
Black, E.; Raghavan, M.; and Barocas, S. 2022.
\newblock Model Multiplicity: Opportunities, Concerns, and Solutions.
\newblock In \emph{Proceedings of the 2022 ACM Conference on Fairness, Accountability, and Transparency}, FAccT '22, 850–863. New York, NY, USA: Association for Computing Machinery.
\newblock ISBN 9781450393522.

\bibitem[{Breiman(2001)}]{breiman_two_cultures}
Breiman, L. 2001.
\newblock {Statistical Modeling: The Two Cultures (with comments and a rejoinder by the author)}.
\newblock \emph{Statistical Science}, 16(3): 199 -- 231.

\bibitem[{Brunet, Anderson, and Zemel(2022)}]{indeterminacy_explanations_zemel}
Brunet, M.-E.; Anderson, A.; and Zemel, R. 2022.
\newblock Implications of Model Indeterminacy for Explanations of Automated Decisions.
\newblock In Koyejo, S.; Mohamed, S.; Agarwal, A.; Belgrave, D.; Cho, K.; and Oh, A., eds., \emph{Advances in Neural Information Processing Systems}, volume~35, 7810--7823. Curran Associates, Inc.

\bibitem[{Cavus and Biecek(2024)}]{cavus2024investigating}
Cavus, M.; and Biecek, P. 2024.
\newblock Investigating the Impact of Balancing, Filtering, and Complexity on Predictive Multiplicity: A Data-Centric Perspective.
\newblock \emph{arXiv preprint arXiv:2412.09712}.

\bibitem[{Chen, Johansson, and Sontag(2018)}]{chen2018my}
Chen, I.; Johansson, F.~D.; and Sontag, D. 2018.
\newblock Why is my classifier discriminatory?
\newblock \emph{Advances in neural information processing systems}, 31.

\bibitem[{Chouldechova(2017)}]{chouldechova_impossibility}
Chouldechova, A. 2017.
\newblock Fair Prediction with Disparate Impact: {A} Study of Bias in Recidivism Prediction Instruments.
\newblock \emph{Big Data}, 5(2): 153--163.

\bibitem[{Cooper et~al.(2023)Cooper, Lee, Choksi, Barocas, Sa, Grimmelmann, Kleinberg, Sen, and Zhang}]{cooper2023arbitrariness}
Cooper, A.~F.; Lee, K.; Choksi, M.; Barocas, S.; Sa, C.~D.; Grimmelmann, J.; Kleinberg, J.; Sen, S.; and Zhang, B. 2023.
\newblock Arbitrariness and Prediction: The Confounding Role of Variance in Fair Classification.
\newblock arXiv:2301.11562.

\bibitem[{Corbett-Davies et~al.(2023)Corbett-Davies, Gaebler, Nilforoshan, Shroff, and Goel}]{corbett2023measure}
Corbett-Davies, S.; Gaebler, J.~D.; Nilforoshan, H.; Shroff, R.; and Goel, S. 2023.
\newblock The measure and mismeasure of fairness.
\newblock \emph{The Journal of Machine Learning Research}, 24(1): 14730--14846.

\bibitem[{Cortez and Silva(2008)}]{cortez2008using}
Cortez, P.; and Silva, A. M.~G. 2008.
\newblock Using data mining to predict secondary school student performance.

\bibitem[{Coston et~al.(2023)Coston, Kawakami, Zhu, Holstein, and Heidari}]{coston2023validity}
Coston, A.; Kawakami, A.; Zhu, H.; Holstein, K.; and Heidari, H. 2023.
\newblock A validity perspective on evaluating the justified use of data-driven decision-making algorithms.
\newblock In \emph{2023 IEEE Conference on Secure and Trustworthy Machine Learning (SaTML)}, 690--704. IEEE.

\bibitem[{Creel and Hellman(2022)}]{creel2022algorithmic}
Creel, K.; and Hellman, D. 2022.
\newblock The algorithmic leviathan: Arbitrariness, fairness, and opportunity in algorithmic decision-making systems.
\newblock \emph{Canadian Journal of Philosophy}, 52(1): 26--43.

\bibitem[{D'Amour et~al.(2022)D'Amour, Heller, Moldovan, Adlam, Alipanahi, Beutel, Chen, Deaton, Eisenstein, Hoffman et~al.}]{d2022underspecification}
D'Amour, A.; Heller, K.; Moldovan, D.; Adlam, B.; Alipanahi, B.; Beutel, A.; Chen, C.; Deaton, J.; Eisenstein, J.; Hoffman, M.~D.; et~al. 2022.
\newblock Underspecification presents challenges for credibility in modern machine learning.
\newblock \emph{The Journal of Machine Learning Research}, 23(1): 10237--10297.

\bibitem[{Darling and Stracuzzi(2018{\natexlab{a}})}]{darling2018toward}
Darling, M.~C.; and Stracuzzi, D.~J. 2018{\natexlab{a}}.
\newblock Toward uncertainty quantification for supervised classification.
\newblock Technical report, Sandia National Lab.(SNL-NM), Albuquerque, NM (United States).

\bibitem[{Darling and Stracuzzi(2018{\natexlab{b}})}]{Darling2018TowardUQ}
Darling, M.~C.; and Stracuzzi, D.~J. 2018{\natexlab{b}}.
\newblock Toward Uncertainty Quantification for Supervised Classification.

\bibitem[{Depeweg et~al.(2018)Depeweg, Hernandez-Lobato, Doshi-Velez, and Udluft}]{depeweg2018decomposition}
Depeweg, S.; Hernandez-Lobato, J.-M.; Doshi-Velez, F.; and Udluft, S. 2018.
\newblock Decomposition of uncertainty in Bayesian deep learning for efficient and risk-sensitive learning.
\newblock In \emph{International conference on machine learning}, 1184--1193. PMLR.

\bibitem[{Der~Kiureghian and Ditlevsen(2009)}]{der2009aleatory}
Der~Kiureghian, A.; and Ditlevsen, O. 2009.
\newblock Aleatory or epistemic? Does it matter?
\newblock \emph{Structural safety}, 31(2): 105--112.

\bibitem[{Ding et~al.(2021)Ding, Hardt, Miller, and Schmidt}]{folktables_ding2021}
Ding, F.; Hardt, M.; Miller, J.; and Schmidt, L. 2021.
\newblock Retiring adult: New datasets for fair machine learning.
\newblock 34: 6478--6490.

\bibitem[{Domingos(2000)}]{unified_decomposition}
Domingos, P. 2000.
\newblock A Unifeid Bias-Variance Decomposition and its Applications.
\newblock 231--238.

\bibitem[{Efron and Tibshirani(1994)}]{efron1994bootstrap}
Efron, B.; and Tibshirani, R.~J. 1994.
\newblock \emph{An introduction to the bootstrap}.
\newblock CRC press.

\bibitem[{Feldman et~al.(2015)Feldman, Friedler, Moeller, Scheidegger, and Venkatasubramanian}]{feldman2015certifying}
Feldman, M.; Friedler, S.~A.; Moeller, J.; Scheidegger, C.; and Venkatasubramanian, S. 2015.
\newblock Certifying and removing disparate impact.
\newblock In \emph{proceedings of the 21th ACM SIGKDD international conference on knowledge discovery and data mining}, 259--268.

\bibitem[{Gal and Ghahramani(2016)}]{pmlr-v48-gal16}
Gal, Y.; and Ghahramani, Z. 2016.
\newblock Dropout as a Bayesian Approximation: Representing Model Uncertainty in Deep Learning.
\newblock In Balcan, M.~F.; and Weinberger, K.~Q., eds., \emph{Proceedings of The 33rd International Conference on Machine Learning}, volume~48 of \emph{Proceedings of Machine Learning Research}, 1050--1059. New York, New York, USA: PMLR.

\bibitem[{Gal et~al.(2022)Gal, Koumoutsakos, Lanusse, Louppe, and Papadimitriou}]{gal2022bayesian}
Gal, Y.; Koumoutsakos, P.; Lanusse, F.; Louppe, G.; and Papadimitriou, C. 2022.
\newblock Bayesian uncertainty quantification for machine-learned models in physics.
\newblock \emph{Nature Reviews Physics}, 4(9): 573--577.

\bibitem[{Ganesh, Taik, and Farnadi(2025)}]{ganesh2025curious}
Ganesh, P.; Taik, A.; and Farnadi, G. 2025.
\newblock The Curious Case of Arbitrariness in Machine Learning.
\newblock \emph{arXiv preprint arXiv:2501.14959}.

\bibitem[{Gebru et~al.(2021)Gebru, Morgenstern, Vecchione, Vaughan, Wallach, Iii, and Crawford}]{gebru2021datasheets}
Gebru, T.; Morgenstern, J.; Vecchione, B.; Vaughan, J.~W.; Wallach, H.; Iii, H.~D.; and Crawford, K. 2021.
\newblock Datasheets for datasets.
\newblock \emph{Communications of the ACM}, 64(12): 86--92.

\bibitem[{Gomez et~al.(2024)Gomez, Machado, Paes, and Calmon}]{gomez2024algorithmic}
Gomez, J.~F.; Machado, C.; Paes, L.~M.; and Calmon, F. 2024.
\newblock Algorithmic arbitrariness in content moderation.
\newblock In \emph{Proceedings of the 2024 ACM Conference on Fairness, Accountability, and Transparency}, 2234--2253.

\bibitem[{Hardt, Price, and Srebro(2016{\natexlab{a}})}]{hardt2016equality}
Hardt, M.; Price, E.; and Srebro, N. 2016{\natexlab{a}}.
\newblock Equality of opportunity in supervised learning.
\newblock \emph{Advances in neural information processing systems}, 29.

\bibitem[{Hardt, Price, and Srebro(2016{\natexlab{b}})}]{hardt_EOP2016}
Hardt, M.; Price, E.; and Srebro, N. 2016{\natexlab{b}}.
\newblock Equality of Opportunity in Supervised Learning.
\newblock In Lee, D.~D.; Sugiyama, M.; von Luxburg, U.; Guyon, I.; and Garnett, R., eds., \emph{Advances in Neural Information Processing Systems 29: Annual Conference on Neural Information Processing Systems 2016, December 5-10, 2016, Barcelona, Spain}, 3315--3323.

\bibitem[{Herasymuk, Arif~Khan, and Stoyanovich(2024)}]{herasymuk2024responsible}
Herasymuk, D.; Arif~Khan, F.; and Stoyanovich, J. 2024.
\newblock Responsible Model Selection with Virny and VirnyView.
\newblock In \emph{Companion of the 2024 International Conference on Management of Data}, 488--491.

\bibitem[{Hofmann(1994)}]{misc_statlog_(german_credit_data)_144}
Hofmann, H. 1994.
\newblock {Statlog (German Credit Data)}.
\newblock UCI Machine Learning Repository.
\newblock {DOI}: https://doi.org/10.24432/C5NC77.

\bibitem[{Holland et~al.(2020)Holland, Hosny, Newman, Joseph, and Chmielinski}]{holland2020dataset}
Holland, S.; Hosny, A.; Newman, S.; Joseph, J.; and Chmielinski, K. 2020.
\newblock The dataset nutrition label.
\newblock \emph{Data Protection and Privacy}, 12(12): 1.

\bibitem[{Hsu and Calmon(2022)}]{hsu2022rashomon}
Hsu, H.; and Calmon, F. 2022.
\newblock Rashomon capacity: A metric for predictive multiplicity in classification.
\newblock \emph{Advances in Neural Information Processing Systems}, 35: 28988--29000.

\bibitem[{H{\"u}llermeier and Waegeman(2021)}]{hullermeier2021aleatoric}
H{\"u}llermeier, E.; and Waegeman, W. 2021.
\newblock Aleatoric and epistemic uncertainty in machine learning: An introduction to concepts and methods.
\newblock \emph{Machine learning}, 110(3): 457--506.

\bibitem[{Jiang et~al.(2023)Jiang, Rago, Leofante, and Toni}]{jiang2023recourse_multiplicity}
Jiang, J.; Rago, A.; Leofante, F.; and Toni, F. 2023.
\newblock Recourse under model multiplicity via argumentative ensembling.
\newblock \emph{arXiv preprint arXiv:2312.15097}.

\bibitem[{Joseph and Raj(2022)}]{joseph2022gandalf}
Joseph, M.; and Raj, H. 2022.
\newblock GANDALF: gated adaptive network for deep automated learning of features.
\newblock \emph{arXiv preprint arXiv:2207.08548}.

\bibitem[{Kamiran, Karim, and Zhang(2012)}]{kamiran2012decision}
Kamiran, F.; Karim, A.; and Zhang, X. 2012.
\newblock Decision theory for discrimination-aware classification.
\newblock In \emph{2012 IEEE 12th international conference on data mining}, 924--929. IEEE.

\bibitem[{Kendall and Gal(2017)}]{gal_vision_neurips}
Kendall, A.; and Gal, Y. 2017.
\newblock What Uncertainties Do We Need in Bayesian Deep Learning for Computer Vision?
\newblock In Guyon, I.; Luxburg, U.~V.; Bengio, S.; Wallach, H.; Fergus, R.; Vishwanathan, S.; and Garnett, R., eds., \emph{Advances in Neural Information Processing Systems}, volume~30. Curran Associates, Inc.

\bibitem[{Khan, Herasymuk, and Stoyanovich(2023)}]{khan2023fairness}
Khan, F.~A.; Herasymuk, D.; and Stoyanovich, J. 2023.
\newblock On Fairness and Stability: Is Estimator Variance a Friend or a Foe?
\newblock \emph{arXiv preprint arXiv:2302.04525}.

\bibitem[{Kleinberg, Mullainathan, and Raghavan(2017)}]{Kleinberg_impossibility}
Kleinberg, J.~M.; Mullainathan, S.; and Raghavan, M. 2017.
\newblock Inherent Trade-Offs in the Fair Determination of Risk Scores.
\newblock In Papadimitriou, C.~H., ed., \emph{8th Innovations in Theoretical Computer Science Conference, {ITCS} 2017, January 9-11, 2017, Berkeley, CA, {USA}}, volume~67 of \emph{LIPIcs}, 43:1--43:23. Schloss Dagstuhl - Leibniz-Zentrum f{\"{u}}r Informatik.

\bibitem[{Lakshminarayanan, Pritzel, and Blundell(2017)}]{lakshminarayanan2017simple}
Lakshminarayanan, B.; Pritzel, A.; and Blundell, C. 2017.
\newblock Simple and scalable predictive uncertainty estimation using deep ensembles.
\newblock \emph{Advances in neural information processing systems}, 30.

\bibitem[{Le~Quy et~al.(2022)Le~Quy, Roy, Iosifidis, Zhang, and Ntoutsi}]{le2022survey}
Le~Quy, T.; Roy, A.; Iosifidis, V.; Zhang, W.; and Ntoutsi, E. 2022.
\newblock A survey on datasets for fairness-aware machine learning.
\newblock \emph{Wiley Interdisciplinary Reviews: Data Mining and Knowledge Discovery}, 12(3): e1452.

\bibitem[{Liu et~al.(2022)Liu, Patwardhan, Grasch, Agarwal et~al.}]{liu2022jitter}
Liu, H.; Patwardhan, S.; Grasch, P.; Agarwal, S.; et~al. 2022.
\newblock Model Stability with Continuous Data Updates.
\newblock \emph{arXiv preprint arXiv:2201.05692}.

\bibitem[{Long et~al.(2023{\natexlab{a}})Long, Hsu, Alghamdi, and Calmon}]{long2023arbitrariness}
Long, C.~X.; Hsu, H.; Alghamdi, W.; and Calmon, F.~P. 2023{\natexlab{a}}.
\newblock Arbitrariness Lies Beyond the Fairness-Accuracy Frontier.
\newblock \emph{arXiv preprint arXiv:2306.09425}.

\bibitem[{Long et~al.(2023{\natexlab{b}})Long, Hsu, Alghamdi, and Calmon}]{individual_arbitrariness}
Long, C.~X.; Hsu, H.; Alghamdi, W.; and Calmon, F.~P. 2023{\natexlab{b}}.
\newblock Individual arbitrariness and group fairness.
\newblock In \emph{Proceedings of the 37th International Conference on Neural Information Processing Systems}, NIPS '23. Red Hook, NY, USA: Curran Associates Inc.

\bibitem[{Malinin(2019)}]{malinin2019uncertainty}
Malinin, A. 2019.
\newblock \emph{Uncertainty estimation in deep learning with application to spoken language assessment}.
\newblock Ph.D. thesis.

\bibitem[{Marx, Calmon, and Ustun(2020)}]{marx2020predictive}
Marx, C.; Calmon, F.; and Ustun, B. 2020.
\newblock Predictive multiplicity in classification.
\newblock In \emph{International Conference on Machine Learning}, 6765--6774. PMLR.

\bibitem[{Meyer, Albarghouthi, and D'Antoni(2023)}]{meyer2023dataset}
Meyer, A.~P.; Albarghouthi, A.; and D'Antoni, L. 2023.
\newblock The dataset multiplicity problem: How unreliable data impacts predictions.
\newblock In \emph{Proceedings of the 2023 ACM Conference on Fairness, Accountability, and Transparency}, 193--204.

\bibitem[{Mitchell et~al.(2019)Mitchell, Wu, Zaldivar, Barnes, Vasserman, Hutchinson, Spitzer, Raji, and Gebru}]{mitchell2019model}
Mitchell, M.; Wu, S.; Zaldivar, A.; Barnes, P.; Vasserman, L.; Hutchinson, B.; Spitzer, E.; Raji, I.~D.; and Gebru, T. 2019.
\newblock Model cards for model reporting.
\newblock In \emph{Proceedings of the conference on fairness, accountability, and transparency}, 220--229.

\bibitem[{Moro, Cortez, and Rita(2014)}]{Moro2014ADA}
Moro, S.; Cortez, P.; and Rita, P. 2014.
\newblock A data-driven approach to predict the success of bank telemarketing.
\newblock \emph{Decis. Support Syst.}, 62: 22--31.

\bibitem[{Mukhoti et~al.(2021)Mukhoti, Kirsch, van Amersfoort, Torr, and Gal}]{mukhoti2021deterministic}
Mukhoti, J.; Kirsch, A.; van Amersfoort, J.; Torr, P.; and Gal, Y. 2021.
\newblock Deterministic neural networks with inductive biases capture epistemic and aleatoric uncertainty.
\newblock \emph{arXiv preprint arXiv:2102.11582}, 2.

\bibitem[{Mukhoti et~al.(2023)Mukhoti, Kirsch, van Amersfoort, Torr, and Gal}]{Mukhoti_2023_CVPR}
Mukhoti, J.; Kirsch, A.; van Amersfoort, J.; Torr, P.~H.; and Gal, Y. 2023.
\newblock Deep Deterministic Uncertainty: A New Simple Baseline.
\newblock In \emph{Proceedings of the IEEE/CVF Conference on Computer Vision and Pattern Recognition (CVPR)}, 24384--24394.

\bibitem[{Nado et~al.(2021)Nado, Band, Collier, Djolonga, Dusenberry, Farquhar, Feng, Filos, Havasi, Jenatton et~al.}]{nado2021uncertainty}
Nado, Z.; Band, N.; Collier, M.; Djolonga, J.; Dusenberry, M.~W.; Farquhar, S.; Feng, Q.; Filos, A.; Havasi, M.; Jenatton, R.; et~al. 2021.
\newblock Uncertainty baselines: Benchmarks for uncertainty \& robustness in deep learning.
\newblock \emph{arXiv preprint arXiv:2106.04015}.

\bibitem[{Paes et~al.(2023)Paes, Cruz, Calmon, and Diaz}]{paes2023inevitability}
Paes, L.~M.; Cruz, R.; Calmon, F.~P.; and Diaz, M. 2023.
\newblock On the inevitability of the Rashomon effect.
\newblock In \emph{2023 IEEE International Symposium on Information Theory (ISIT)}, 549--554. IEEE.

\bibitem[{Pleiss et~al.(2017)Pleiss, Raghavan, Wu, Kleinberg, and Weinberger}]{pleiss2017fairness}
Pleiss, G.; Raghavan, M.; Wu, F.; Kleinberg, J.; and Weinberger, K.~Q. 2017.
\newblock On fairness and calibration.
\newblock \emph{Advances in neural information processing systems}, 30.

\bibitem[{Rahaman and thiery(2021)}]{NEURIPS2021_rahaman_UQ}
Rahaman, R.; and thiery, a. 2021.
\newblock Uncertainty Quantification and Deep Ensembles.
\newblock In Ranzato, M.; Beygelzimer, A.; Dauphin, Y.; Liang, P.; and Vaughan, J.~W., eds., \emph{Advances in Neural Information Processing Systems}, volume~34, 20063--20075. Curran Associates, Inc.

\bibitem[{Simson, Pfisterer, and Kern(2024)}]{simson2024one}
Simson, J.; Pfisterer, F.; and Kern, C. 2024.
\newblock One model many scores: using multiverse analysis to prevent fairness hacking and evaluate the influence of model design decisions.
\newblock In \emph{Proceedings of the 2024 ACM Conference on Fairness, Accountability, and Transparency}, 1305--1320.

\bibitem[{Stoyanovich and Howe(2019)}]{stoyanovich2019nutritional}
Stoyanovich, J.; and Howe, B. 2019.
\newblock Nutritional labels for data and models.
\newblock \emph{A Quarterly bulletin of the Computer Society of the IEEE Technical Committee on Data Engineering}, 42(3).

\bibitem[{Subbaswamy, Adams, and Saria(2021)}]{subbaswamy2021evaluating}
Subbaswamy, A.; Adams, R.; and Saria, S. 2021.
\newblock Evaluating model robustness and stability to dataset shift.
\newblock In \emph{International Conference on Artificial Intelligence and Statistics}, 2611--2619. PMLR.

\bibitem[{Tahir, Cheng, and Liu(2023)}]{tahir_aleatoric_Fairness}
Tahir, A.; Cheng, L.; and Liu, H. 2023.
\newblock Fairness through Aleatoric Uncertainty.
\newblock In \emph{Proceedings of the 32nd ACM International Conference on Information and Knowledge Management}, CIKM '23, 2372–2381. New York, NY, USA: Association for Computing Machinery.
\newblock ISBN 9798400701245.

\bibitem[{Tigga and Garg(2020)}]{diabetes_dataset}
Tigga, N.~P.; and Garg, S. 2020.
\newblock Prediction of Type 2 Diabetes using Machine Learning Classification Methods.
\newblock \emph{Procedia Computer Science}, 167: 706--716.
\newblock International Conference on Computational Intelligence and Data Science.

\bibitem[{Wang et~al.(2024)Wang, He, Gao, and Calmon}]{wang2024aleatoric}
Wang, H.; He, L.; Gao, R.; and Calmon, F. 2024.
\newblock Aleatoric and epistemic discrimination: Fundamental limits of fairness interventions.
\newblock \emph{Advances in Neural Information Processing Systems}, 36.

\bibitem[{Watson-Daniels, Parkes, and Ustun(2023{\natexlab{a}})}]{Watson_Daniels_Parkes_Ustun_2023}
Watson-Daniels, J.; Parkes, D.~C.; and Ustun, B. 2023{\natexlab{a}}.
\newblock Predictive Multiplicity in Probabilistic Classification.
\newblock \emph{Proceedings of the AAAI Conference on Artificial Intelligence}, 37(9): 10306--10314.

\bibitem[{Watson-Daniels, Parkes, and Ustun(2023{\natexlab{b}})}]{watson2023predictive}
Watson-Daniels, J.; Parkes, D.~C.; and Ustun, B. 2023{\natexlab{b}}.
\newblock Predictive multiplicity in probabilistic classification.
\newblock In \emph{Proceedings of the AAAI Conference on Artificial Intelligence}, volume~37, 10306--10314.

\bibitem[{Wightman(1998)}]{wightman1998lsac}
Wightman, L.~F. 1998.
\newblock LSAC National Longitudinal Bar Passage Study. LSAC Research Report Series.

\bibitem[{Xiao, Gomez, and Gal(2019)}]{xiao2019wat}
Xiao, T.~Z.; Gomez, A.~N.; and Gal, Y. 2019.
\newblock Wat heb je gezegd? detecting out-of-distribution translations with variational transformers.
\newblock In \emph{Bayesian Deep Learning Workshop (NeurIPS)}.

\bibitem[{Zemel et~al.(2013)Zemel, Wu, Swersky, Pitassi, and Dwork}]{zemel2013learning}
Zemel, R.; Wu, Y.; Swersky, K.; Pitassi, T.; and Dwork, C. 2013.
\newblock Learning fair representations.
\newblock In \emph{International conference on machine learning}, 325--333. PMLR.

\bibitem[{Zhang, Lemoine, and Mitchell(2018)}]{zhang2018mitigating}
Zhang, B.~H.; Lemoine, B.; and Mitchell, M. 2018.
\newblock Mitigating unwanted biases with adversarial learning.
\newblock In \emph{Proceedings of the 2018 AAAI/ACM Conference on AI, Ethics, and Society}, 335--340.

\bibitem[{Zhang, Kailkhura, and Han(2020)}]{zhang2020mix}
Zhang, J.; Kailkhura, B.; and Han, T. Y.-J. 2020.
\newblock Mix-n-match: Ensemble and compositional methods for uncertainty calibration in deep learning.
\newblock In \emph{International conference on machine learning}, 11117--11128. PMLR.

\end{thebibliography}

\appendix
\clearpage
\section{Appendix}
\label{sec:appendix}

\subsection{Dataset Descriptions and Statistics}
\label{sec:dataset_info_appdx}
\begin{itemize}
    \item \textbf{Folktables}\footnote{ \url{https://github.com/zykls/folktables}} \cite{folktables_ding2021}. This benchmark is a suite of modern datasets derived from US Census surveys from 50 US states for the years 2014-2018. It is associated with several prediction tasks related to income, healthcare, transportation, employment, and housing. We select two of them (listed below). For our group fairness analysis, we construct groups based on two sensitive attributes, namely $sex$, which is binarized into ''female'' (disadvantaged) and ''male'' (privileged) groups and $race$, which we binarize into ''Non-white'' (disadvantaged) and ''White'' (privileged). For our study, we report disparity metrics on intersectional groups, ie. ''non-white, female'' (intersectionally disadvantaged) vs all other samples. 
    
    \begin{itemize}
        \item \textbf{ACS Income}. This is a binary classification task to predict whether a person's income is $< \$50,000$ (label 0) or $>= \$50,000$ (label 1). This dataset has 8 categorical and 2 numerical features, including education, occupation, and marital status. For our study, we used data from Georgia, 2018, and sub-sampled the data from 50,915 to 15,000 rows for computational feasibility.
        \item \textbf{ACS Public Coverage}. This is a binary classification task to predict whether an individual is on public health insurance (label 1) or not (label 0). This dataset contains 17 categorical and 2 numerical features, including education, disability, and marital status. For our study, we used data from California in 2018, and sub-sample the data from 138,554 to 15,000 rows for computational feasibility.
    \end{itemize}
    
    \item \textbf{Law School}\footnote{ \url{https://github.com/tailequy/fairness_dataset/blob/main/experiments/data/law_school_clean.csv}} \cite{wightman1998lsac}. This dataset was gathered through a survey conducted by the Law School Admission Council (LSAC) across 163 law schools in the United States in 1991. It encompasses records of law school admissions. The dataset contains information of 20,798 students characterized by 11 attributes (5 categorical and 6 numerical attributes). The associated prediction task is to predict whether a candidate would pass the bar exam (label 1) or not (label 0). For our group fairness analysis, we construct groups using two sensitive attributes: $sex$ (''male'' for privileged, ''female'' for disadvantaged) and $race$ (''White'' for privileged, ``Non-White'' for disadvantaged). Similar to Folktables, we report disparity metrics on intersectional $sex\&race$ groups.

    \item \textbf{Student Performance}\footnote{ \url{https://github.com/tailequy/fairness_dataset/blob/main/experiments/data/student_por_new.csv}} \cite{cortez2008using}. This dataset provides information on the academic performance of students in the secondary education of two Portuguese schools in 2005 - 2006 with two distinct subjects: Mathematics and Portuguese. For our experiments, we selected a dataset related to the Portuguese subject due to its larger dataset size. It contains information of 649 students characterized by 33 attributes (4 categorical, 13 binary and 16 numerical attributes). The initial regression task was to predict the final year grade of the students. To get a binary classification task, we used a preprocessed dataset from \cite{le2022survey}. The target label is derived from the attribute \textit{G3} (representing the final grade), where \textit{target} = \{Low, High\}, corresponding to \textit{G3} = \{<10, $\geq$10\}. The positive class is denoted as "High." There is only one traditional sensitive attribute $sex$ = \{female, male\}, which we used for our study.

    \item \textbf{\german}\footnote{\url{https://archive.ics.uci.edu/dataset/144/statlog+german+credit+data}}~\cite{misc_statlog_(german_credit_data)_144} is a popular fairness dataset that contains records of creditworthiness assessments, classifying individuals as high or low credit risks. It contains information on 1,000 individuals characterized by 21 attributes (14 categorical, 7 numerical), including credit history, occupation and housing information. Here, \textit{sex} and \textit{age} are the sensitive attributes, with ``female'' and ``age$\leq25$'' as the disadvantaged groups.

    \item \textbf{\bank}\footnote{\url{https://archive.ics.uci.edu/dataset/222/bank+marketing}}
~\cite{Moro2014ADA} contains data from direct marketing campaigns by a Portuguese bank, between 2008 and 2013. It contains information on 40,004 potential customers, with 16 attributes (10 categorical, 6 numerical), including occupation, marital status, education, and a binary target that indicates whether the individual subscribed for a term deposit. Here, \textit{age} is the sensitive attribute, with $<25$ and $>60$ as the disadvantaged group. 

    \item \textbf{\diabetes}\footnote{\url{https://www.kaggle.com/datasets/tigganeha4/diabetes-dataset-2019}}~\cite{diabetes_dataset} was collected in India through a questionnaire including 18 questions related to health, lifestyle, and family background. A total of 952 participants are characterized by 17 attributes (13 categorical, 4 numerical) and a binary target variable that represents whether a person is diabetic. Here, \textit{sex} is the sensitive attribute, with ``female'' as the disadvantaged group.
\end{itemize}

\begin{table*}[h!]
\centering
\begin{tabular}{|c|c|c|c|c|c|c|c|}
\hline
 & overall & sex$\_$priv & sex$\_$dis &race$\_$priv & race$\_$dis & sex$\&$race$\_$priv &
sex$\&$race$\_$dis \\
\hline
Proportion & 1.0 & 0.511 & 0.489 & 0.678 & 0.322
& 0.829 & 0.171 \\
Base Rate & 0.35 & 0.422 & 0.274 & 0.389 & 0.267
& 0.374 & 0.232 \\
\hline
    \end{tabular}
    \caption{ACS Income, 2018 year, GA state (number of rows = 15K)}
    \label{tab:income-stats}
\end{table*}

\begin{table*}[h!]
\centering
\begin{tabular}{|c|c|c|c|c|c|c|c|}
\hline
 & overall & sex$\_$priv & sex$\_$dis &race$\_$priv & race$\_$dis & sex$\&$race$\_$priv &
sex$\&$race$\_$dis \\
\hline
Proportion & 1.0 & 0.445 & 0.555 & 0.565 & 0.435 & 0.761 & 0.239 \\
Base Rate & 0.374 & 0.395 & 0.357 & 0.363 & 0.388 & 0.374 & 0.374 \\
\hline
    \end{tabular}
    \caption{ACS Public Coverage, 2018 year, CA state (number of rows = 15K)}
    \label{tab:pubcov-stats}
\end{table*}

\begin{table*}[h!]
\centering
\begin{tabular}{|c|c|c|c|c|c|c|c|}
\hline
& overall & male$\_$priv & male$\_$dis & race$\_$priv & race$\_$dis & male$\&$race$\_$priv &
male$\&$race$\_$dis \\
\hline
Proportion & 1.0 & 0.561 & 0.439 & 0.841 & 0.159 & 0.917 & 0.083 \\
Base Rate & 0.89 & 0.899 & 0.878 & 0.921 & 0.723 & 0.906 & 0.713 \\
\hline
    \end{tabular}
    \caption{Law School (number of rows = 20,798)}
    \label{tab:law-stats}
\end{table*}

\begin{table*}[h!]
\centering
\begin{tabular}{|c|c|c|c|}
\hline
& overall & sex$\_$priv & sex$\_$dis \\
\hline
Proportion & 1.0 & 0.41 & 0.59 \\
Base Rate & 0.846 & 0.812 & 0.869 \\
\hline
    \end{tabular}
    \caption{Student Performance, Portuguese subject (number of rows = 649)}
    \label{tab:student-stats}
\end{table*}

\begin{table*}[h!]
    \centering
    \begin{tabular}{|c|c|c|c|}
    \hline
     & overall & gender\_priv & gender\_dis \\ 
    \hline
    Proportions & 1.0 & 0.621 & 0.379 \\ 
    \hline
    Base Rates & 0.291 & 0.272 & 0.321 \\ 
    \hline
    \end{tabular}
    \caption{Table~\ref{tab:diabetes-rates}: Proportions and Base Rates for \diabetes.}
    \label{tab:diabetes-rates}
\end{table*}

\begin{table*}[h!]
\centering
    \begin{tabular}{|c|c|c|c|c|c|c|c|}
    \hline
     & overall & sex\_priv & sex\_dis & age\_priv & age\_dis & sex\&age\_priv & sex\&age\_dis \\ 
    \hline
    Proportions & 1.0 & 0.69 & 0.31 & 0.81 & 0.19 & 0.895 & 0.105 \\ 
    \hline
    Base Rates & 0.7 & 0.723 & 0.648 & 0.728 & 0.579 & 0.717 & 0.552 \\ 
    \hline
    \end{tabular}
\caption{Table~\ref{tab:german-rates}: Proportions and Base Rates for \german.}
\label{tab:german-rates}
\end{table*}

\begin{table*}[h!]
    \centering
    \begin{tabular}{|c|c|c|c|}
    \hline
     & overall & age\_priv & age\_dis \\ 
    \hline
    Proportions & 1.0 & 0.955 & 0.045 \\ 
    \hline
    Base Rates & 0.117 & 0.106 & 0.341 \\ 
    \hline
    \end{tabular}
    \caption{Table~\ref{tab:bank-rates}: Proportions and Base Rates for \bank.}
    \label{tab:bank-rates}
\end{table*}

\begin{table*}[h!]
\centering
\begin{tabular}{|c|c|c|c|c|c|c|c|}
\hline
 & overall & sex$\_$priv & sex$\_$dis &race$\_$priv & race$\_$dis & sex$\&$race$\_$priv &
sex$\&$race$\_$dis \\
\hline
Proportion & 1.0 & 0.509 & 0.491 & 0.737 & 0.263
& 0.866 & 0.134 \\
Base Rate & 0.482 & 0.551 & 0.411 & 0.513 & 0.395
& 0.501 & 0.359 \\
\hline
    \end{tabular}
    \caption{ACS Income, 2018 year, \{MD, NJ, MA\} states (high-income dataset, number of rows = 100K)}
    \label{tab:high-income-stats}
\end{table*}

\begin{table*}[h!]
\centering
\begin{tabular}{|c|c|c|c|c|c|c|c|}
\hline
 & overall & sex$\_$priv & sex$\_$dis &race$\_$priv & race$\_$dis & sex$\&$race$\_$priv &
sex$\&$race$\_$dis \\
\hline
Proportion & 1.0 & 0.516 & 0.484 & 0.788 & 0.212 & 0.887 & 0.113 \\
Base Rate & 0.302 & 0.386 & 0.212 & 0.333 & 0.186 & 0.322 & 0.14 \\
\hline
    \end{tabular}
    \caption{ACS Income, 2018 year, \{WV, MS, AR, NM, LA, AL, KY\} states (low-income dataset, number of rows = 100K)}
    \label{tab:low-income-stats}
\end{table*}

\subsection{Model Tuning}
\label{sec:model_info_appdx}
We tuned the following hyperparameters for each model type: (i) decision tree (\texttt{dt$\_$clf}) with a tuned maximum tree depth, minimum samples at a leaf node, number of features used to decide the best split, and criteria to measure the quality of a split; (ii) logistic regression (\texttt{lr$\_$clf}) with tuned regularization penalty, regularization strength, and optimization algorithm; (iii) gradient boosted trees (\texttt{lgbm$\_$clf)} with tuned number of boosted trees, maximum tree depth, maximum tree leaves, and minimum number of samples in a leaf; (iv) random forest (\texttt{rf$\_$clf}) with a tuned number of trees, maximum tree depth, minimum samples required to split a node, and minimum samples at a leaf node (v) neural network, historically called the multi-layer perceptron (\texttt{mlp$\_$clf}) with two hidden layers, each with 100 neurons, and a tuned activation function, optimization algorithm, and learning rate; (vi) a deep table-learning method called GANDALF~\cite{joseph2022gandalf} (\texttt{gandalf$\_$clf)} with a tuned learning rate, number of layers in the feature abstraction layer, dropout rate for the feature abstraction layer, and initial percentage of features to be selected in each Gated Feature Learning Unit (GFLU) stage.

\subsection{F1-Stability Trade-offs of Different Model Types}
\label{sec:models-tradeoff-appendix}

\begin{figure}[h]
\begin{subfigure}[h]{\linewidth}
    \centering
    \includegraphics[width=\linewidth]{images/baselines/folk-income-models.png}
    \caption{Folk-Income}
\end{subfigure}
\hfill
\begin{subfigure}[h]{\linewidth}
    \centering
    \includegraphics[width=\linewidth]{images/baselines/law-school-models.png}
    \caption{Law School}
\end{subfigure}
 \hfill
\begin{subfigure}[h]{\linewidth}
    \centering
    \includegraphics[width=\linewidth]{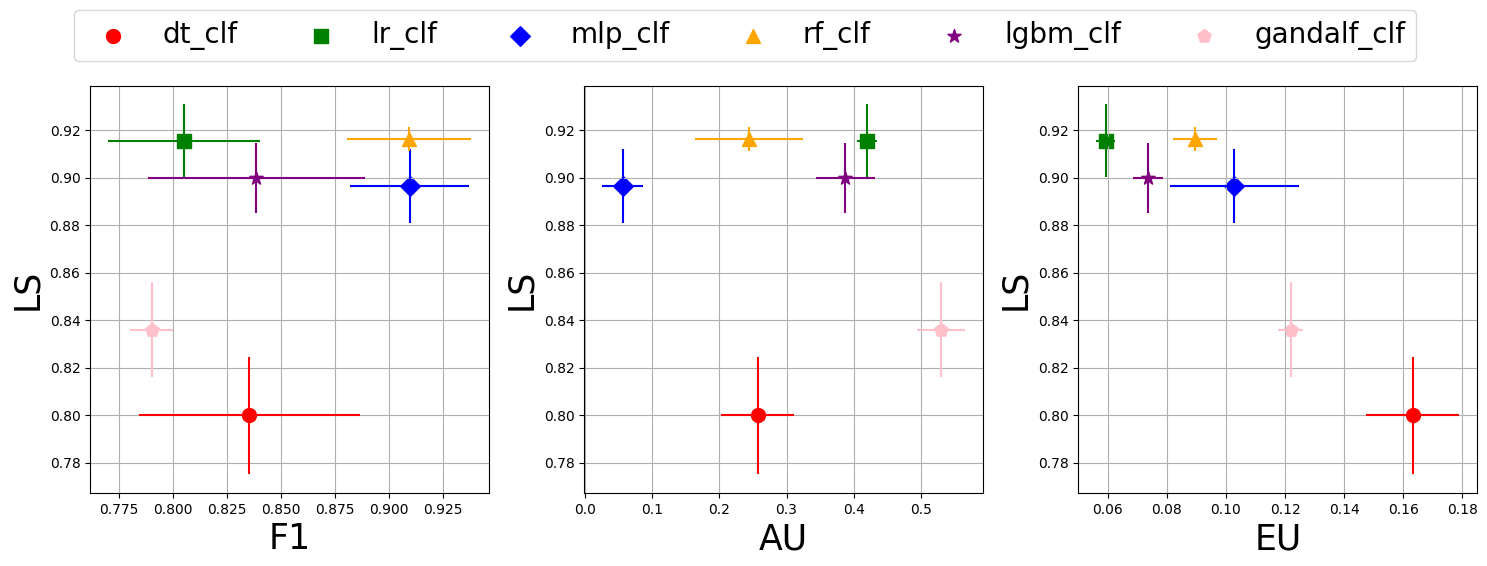}
    \caption{Diabetes}
\end{subfigure}
\hfill
\begin{subfigure}[h]{\linewidth}
    \centering
    \includegraphics[width=\linewidth]{images/baselines/german-models.png}
    \caption{German}
\end{subfigure}
\hfill
\begin{subfigure}[h]{\linewidth}
    \centering
    \includegraphics[width=\linewidth]{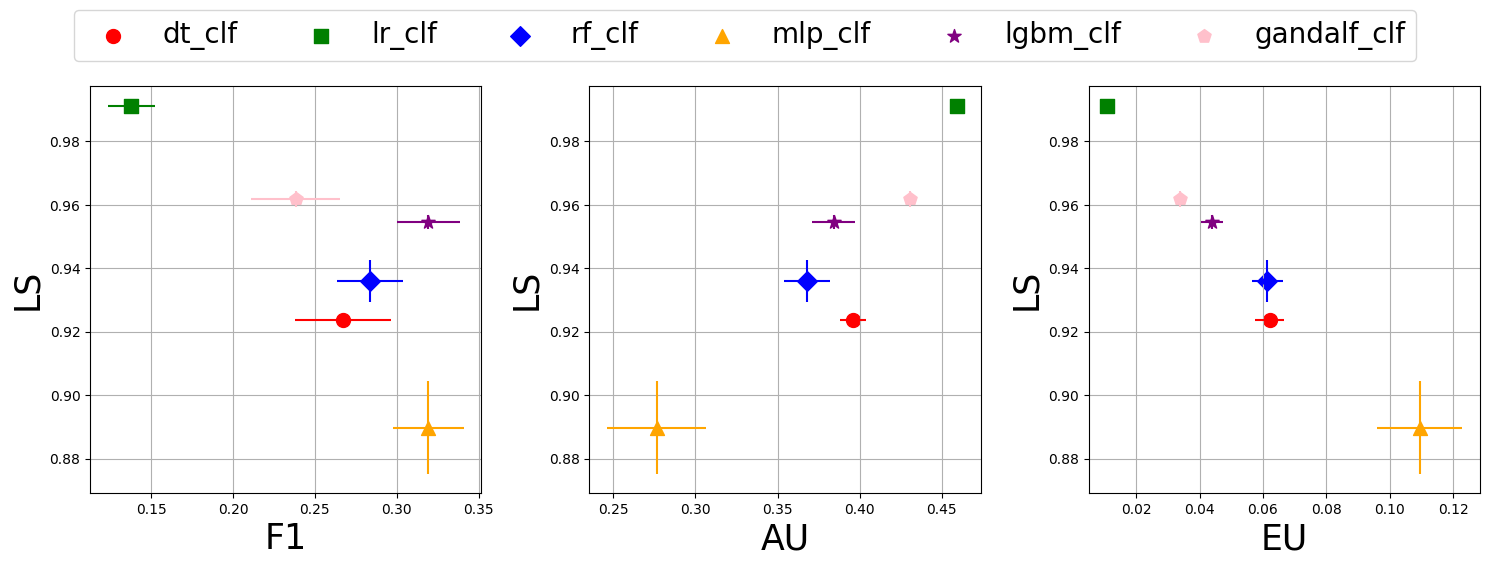}
    \caption{Bank}
\end{subfigure}
\hfill
\caption{Different models make different stability-accuracy trade-offs, and label stability decomposes into epistemic and aleatoric components.}
\label{fig:stability-tradeoffs}
\end{figure}

\subsection{Out-of-domain prediction results on high-income domain}
\label{sec:OOD-appdx}
\begin{figure*}
    \centering
    \includegraphics[width=\linewidth]{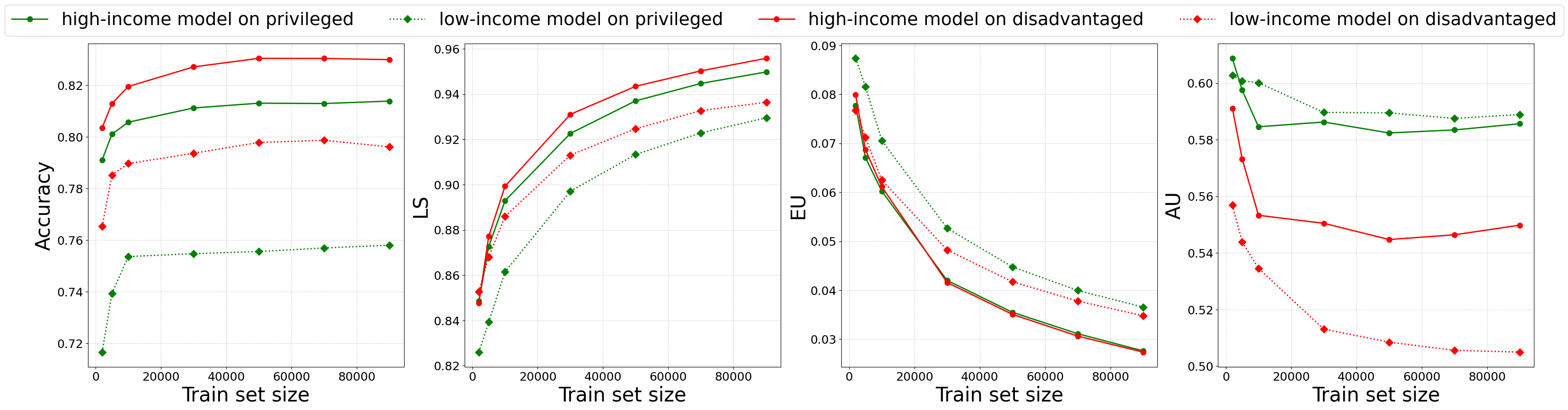}
    \caption{Testing on the high income domain. In-domain models (trained on high income domain) are showed in bold lines, out-domain models (trained on the low income domain) are shown in dashed lines. Average performance on the privileged group is shown in green, and on the disadvantaged in red.}
    \label{fig:high-income-testing}
\end{figure*}

\subsection{The \libname software library}
\label{sec:library-additional-appdx}


As mentioned in Section~\ref{sec:prelims}, the \libname software library decouples the process of model profiling into several stages, namely \textit{subgroup metric computation}, \textit{disparity metric composition}, and \textit{metric visualization}. This separation provides data scientists with enhanced control and flexibility when utilizing the library, both during model development and for post-deployment monitoring. Figure~\ref{fig:library_diagram} illustrates how the library constructs a pipeline for model analysis. Inputs to a user interface are depicted in green, pipeline stages are represented in blue, and the output of each stage is shown in purple. We will now describe each of these stages.

\subsubsection{Inputs}

To use the library, users need to provide three inputs, namely:

\begin{itemize}
    \item A \textit{base flow dataset} is a custom object for the user's dataset that contains its descriptive attributes such as a target column, numerical columns, categorical columns, train and test sets, etc. This object must be inherited from the $BaseFlowDataset$ class, created for user ease. The idea behind having a common base class is to validate the user's input at the beginning of processing and to subsequently simplify the logic for downstream metric computation.

    \item A \textit{config yaml} is a file employed to define configuration parameters for metric computation interfaces within the framework. This user-defined configuration approach enhances flexibility for users. With just one \textit{config yaml} per experiment, users can easily shift between different experiments without the need for additional adjustments when utilizing user interfaces from the library. Furthermore, the config file is a convenient way to define complex configurations for protected groups and consolidate all model auditing parameters within one place.

    The \textit{config yaml} contains information such as the number of estimators in an ensemble for variance analysis, the fraction of samples for random sampling of an input train set, a metric computation mode, etc. Importantly, we request the user to define protected groups from the dataset by simply passing a dictionary where key-value pairs specify the relevant column names and the disadvantaged value or a list of values of the sensitive attribute for partitioning on subgroups. Users can also specify intersectional groups here. 
    
    \item Finally, a \textit{models config} is a Python dictionary that serves as a key-value mapping, where keys are model names and values are initialized models for analysis. This dictionary facilitates conducting audits on multiple models, as well as enables the analysis of various types of models.
\end{itemize}

\subsubsection{Subgroup metric computation} 

The library provides several user interfaces for metric computation: an interface for multiple models, an interface for multiple test sets, and an interface for saving results into a user-defined database. After the variables are input to a user interface, the base flow dataset is used in subgroup analyzers to compute diverse metric sets. Our library incorporates a \textit{Subgroup Variance Analyzer} and a \textit{Subgroup Error Analyzer} (to be compatible with the model error decomposition proposed by \citet{unified_decomposition}), and it is easily extensible to encompass other analyzers. Once these analyzers finalize metric computation, their outputs are merged and returned as a pandas dataframe. Moreover, users have the option to specify a parameter for saving, allowing the metric dataframe to be stored on disk or in their database using the provided \textit{df\_writer} function.

The \textit{Subgroup Variance Analyzer} is responsible for computing our stability and uncertainty metrics on both the overall test set and user-specified protected groups. To quantify estimator variance, we use a bootstrapping approach~\cite{efron1994bootstrap}. However, instead of a straightforward computation of the standard deviation of the predictive distribution, we extend this computation to include additional metrics like Label Stability~\cite{Darling2018TowardUQ}, Jitter~\cite{liu2022jitter}, and Std/IQR (standard deviation / inter-quantile range of predictive variance). Similarly, the \textit{Subgroup Error Analyzer} computes error metrics (such as accuracy, F1, FPR, FNR, etc.) on both the overall test set and the specified subgroups of interest.

\subsubsection{Disparity metric composition} 

The second stage of the model profiling process is managed by the \textit{Metric Composer}, which computes metrics such as error disparity, stability disparity, and uncertainty disparity. Nevertheless, users retain the flexibility to compose additional metrics if needed. For instance, the error disparity measure of Disparate Impact is constructed by deriving the ratio of the Positive Rate computed on the privileged  and disadvantaged subgroups.

\subsubsection{Metric visualization} 

\libname offers two types of metric visualizers designed for creating static and interactive visualizations: the \textit{Metric Visualizer} and the \textit{Metric Interactive Visualizer}. With the \textit{Metric Visualizer}, users can easily produce tailored static visualizations for in-depth metric analysis, covering both overall and disparity metrics. On the other hand, the \textit{Metric Interactive Visualizer} provides users with the capability to construct an interactive web application. This application guides the responsible model selection process and generates nutritional labels for ML models.

\subsection{Fairness interventions}
\label{sec:fairness_interventions_descriptions}

\begin{itemize}
    \item \textbf{Disparate Impact Remover (DIR) \cite{feldman2015certifying}}: A \textit{pre-processing} technique designed to adjust feature values, enhancing group fairness while maintaining rank-ordering within groups.

    \item \textbf{Learning Fair Representations (LFR) \cite{zemel2013learning}}: A \textit{pre-processing} method aimed at discovering a latent representation that effectively encodes the data but obfuscates information related to protected attributes.
    
    \item \textbf{Adversarial Debiasing (ADB) \cite{zhang2018mitigating}}: An \textit{in-processing} technique that trains a classifier to attain high prediction accuracy while concurrently reducing the adversary's ability to deduce protected attributes from the predictions. The outcome is a fair classifier, as the predictions no longer contain information about group discrimination that could be exploited by the adversary.

    \item \textbf{Exponentiated Gradient Reduction (EGR) \cite{agarwal2018reductions}}: An \textit{in-processing} technique that simplifies fair classification into a series of cost-sensitive classification problems. It yields a randomized classifier with the minimal empirical error while adhering to fair classification constraints.

    \item \textbf{Equalized-Odds Postprocessing (EOP) \cite{hardt2016equality, pleiss2017fairness}}: A \textit{post-processing} technique that resolves a linear program to determine probabilities for altering output labels, optimizing equalized odds.

    \item \textbf{Reject Option Classification (ROC) \cite{kamiran2012decision}}: A \textit{post-processing} technique that provides favorable outcomes to unprivileged groups and unfavorable outcomes to privileged groups within a confidence band around the decision boundary with the highest uncertainty.
\end{itemize}

\textbf{Experimental procedure.} A single experimental run proceeds as follows: We first randomly split the dataset into training and test sets (80:20 split). Depending on the type of a fairness intervention, we employ a processor from IBM's AIF 360 toolkit~\cite{bellamy2019ai} in a manner that aligns with the specific intervention. In the case of a fair pre-processor, we fit the pre-processor on the training set and subsequently transform both the training and the test sets. In the case of a fair in-processor, we wrote a wrapper for AIF360 and pass this wrapped in-processor as a basic model to a metric computation interface from \libname. In the case of a fair post-processor, we initialize the post-processor and pass it as an additional parameter to a metric computation interface from \libname, along with models for profiling. \libname has the capability to manage the downstream application of the post-processor from AIF360 during model profiling, streamlining the post-processing procedure for users. All fairness interventions are utilized based on an intersectional \textit{sex\&race} column for all datasets, except for Student Performance, which contains only \textit{sex} sensitive attribute.

Following this, we initialize the base model $H$ and tune hyper-parameters once for each fairness intervention and model-type pair. We then run the bootstrap procedure to construct the approximating ensemble with $m=200$ members, with each model trained using the optimized hyper-parameter settings of the base model $H$. We repeat this procedure 6 times, each with a different random train-test split, for each fairness intervention and model-type pair.

\subsection{Model selection results}

\begin{figure*}[t!]
     \centering
    \begin{subfigure}[b]{0.45\linewidth}
         \centering
         \includegraphics[width=0.9\linewidth]{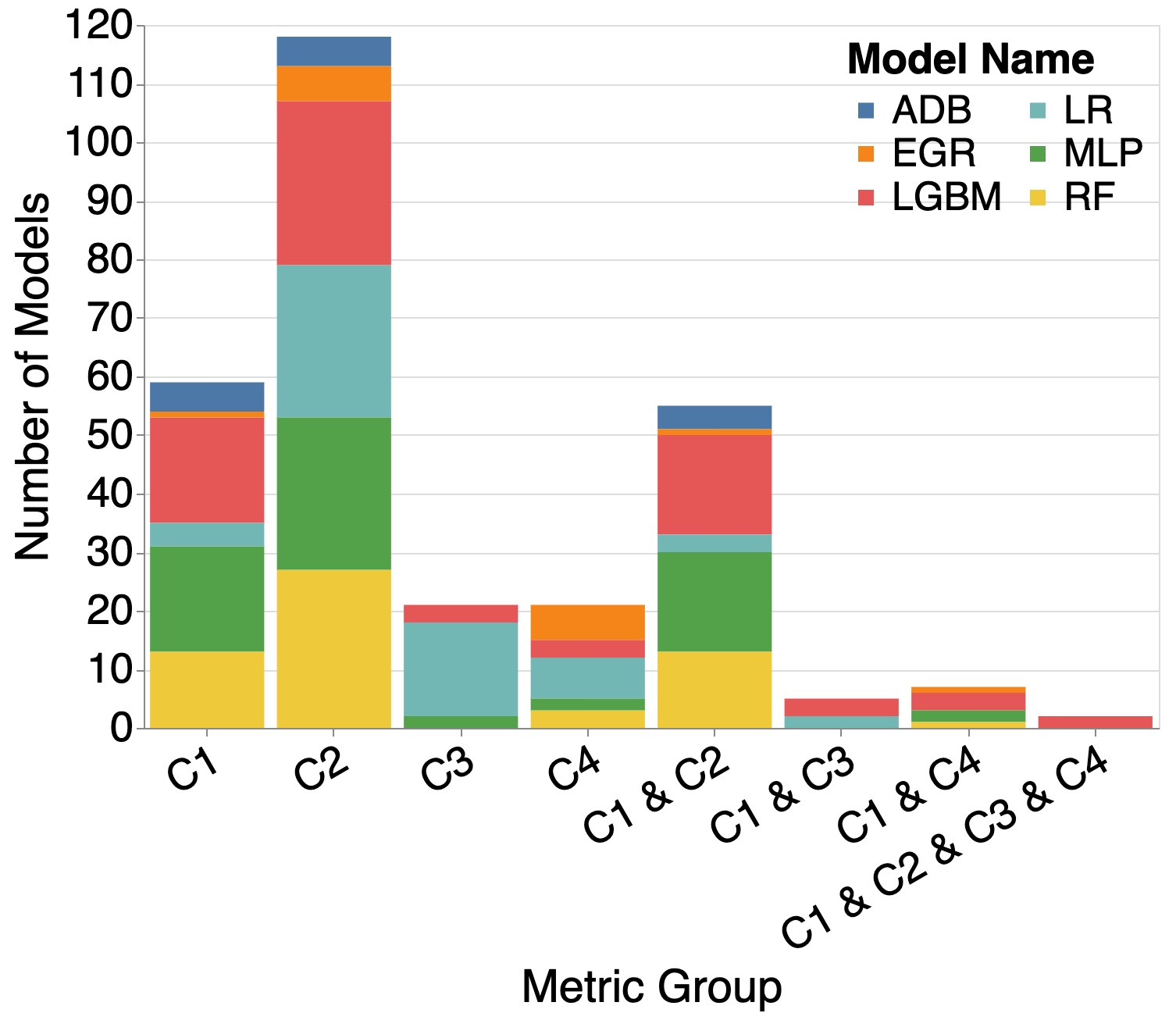}
         \caption{\folkcov: C1: Accuracy $\in$ [0.7, 1], C2: FPRD $\in$ [-0.03, 0.03], C3: $\Delta EU$ $\in$ [-0.001, 0.001], C4: $\Delta AU$ $\in$ [-0.01, 0.01]}
         \label{fig:model-selection-folk-pubcov}
     \end{subfigure}
     \hspace{10pt}
     \hfill
     \begin{subfigure}[b]{0.45\linewidth}
         \centering
         \includegraphics[width=0.9\linewidth]{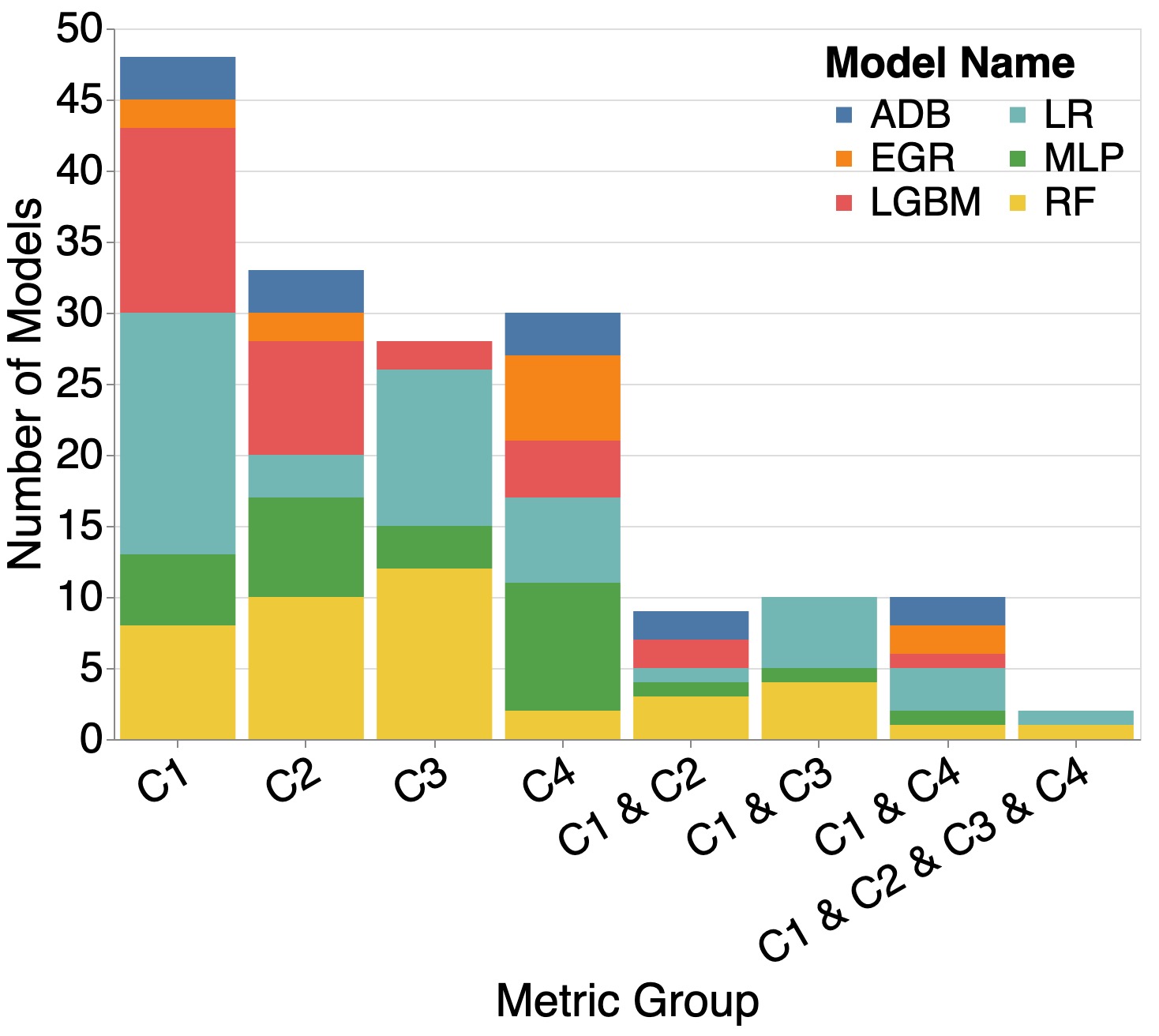}
         \caption{\student: C1: Accuracy $\in$ [0.93, 1], C2: FPRD $\in$ [-0.04, 0.04], C3: $\Delta EU$ $\in$ [-0.01, 0.01], C4: $\Delta AU$ $\in$ [-0.045, 0.045]}
         \label{fig:model-selection-student}
     \end{subfigure}
\caption{Incorporating stability-parity criteria during model selection.}
\label{fig:model-selection-full}
\end{figure*}

\end{document}